\documentclass[lettersize,journal]{IEEEtran}
\usepackage{amsmath,amsfonts}
\usepackage{algorithmic}
\usepackage{algorithm}
\usepackage{array}
\usepackage[caption=false,font=normalsize,labelfont=sf,textfont=sf]{subfig}
\usepackage{textcomp}
\usepackage{stfloats}
\usepackage{url}
\usepackage{verbatim}
\usepackage{graphicx}
\usepackage{cite}
\usepackage{placeins}

\usepackage{amssymb}
\usepackage{amstext}
\usepackage{amsmath}
\usepackage{multicol, multirow}
\usepackage{dsfont}
\usepackage{pifont}
\usepackage[dvipsnames]{xcolor}
\usepackage{overpic}
\usepackage{tikz}
\usepackage{soul}
\usepackage{url}
\usepackage{listings}
\usepackage{lipsum}
\usepackage{placeins}
\usepackage{booktabs,hyphenat} 
\usepackage{cite,hyperref}

\usepackage{adjustbox}

\usepackage[resetlabels,labeled]{multibib}
\newcites{R}{References}
\usepackage[switch]{lineno}

\newcommand{\mname}[0]{\mbox{FusionRAFT}}
\usepackage{soul}
\usepackage{todonotes}
\usepackage{bm}

\begin{document}



\title{Attentive Multimodal Fusion for \\Optical and Scene Flow}

\author{Youjie Zhou$^1$, Guofeng Mei$^2$, Yiming Wang$^2$, Fabio Poiesi$^2$, Yi Wan$^1$
\thanks{This work was supported by the China government project (2019JZZY010112), (2020JMRH0202), (YC-KYXM-07-2021) and by the PNRR project FAIR - Future AI Research (PE00000013), under the NRRP MUR program funded by the NextGenerationEU.}
\thanks{$^{1}$Youjie Zhou and Yi Wan are with the School of Mechanical Engineering, Shandong University, China,
and the Key Laboratory of High Efficiency and Clean Mechanical Manufacture of Ministry of Education, Shandong University, China. {\tt 202020511@mail.sdu.edu.cn,<wanyi>@sdu.edu.cn}. Yi Wan is the corresponding author.}%
\thanks{$^{2}$Guofeng Mei, Yiming Wang, and Fabio Poiesi are with Fondazione Bruno Kessler, Italy {\tt <gmei,ywang,poiesi>@fbk.eu}.}
}

\markboth{IEEE Robotics and Automation Letters. Preprint Version. July, 2023}%
{Zhou \MakeLowercase{\textit{et al.}}: Attentive Multimodal Fusion for Optical and Scene Flow}

\maketitle

\begin{abstract}
This paper presents an investigation into the estimation of optical and scene flow using RGBD information in scenarios where the RGB modality is affected by noise or captured in dark environments. Existing methods typically rely solely on RGB images or fuse the modalities at later stages, which can result in lower accuracy when the RGB information is unreliable. To address this issue, we propose a novel deep neural network approach named {\mname}, which enables early-stage information fusion between sensor modalities (RGB and depth). Our approach incorporates self- and cross-attention layers at different network levels to construct informative features that leverage the strengths of both modalities. Through comparative experiments, we demonstrate that our approach outperforms recent methods in terms of performance on the synthetic dataset Flyingthings3D, as well as the generalization on the real-world dataset KITTI. We illustrate that our approach exhibits improved robustness in the presence of noise and low-lighting conditions that affect the RGB images. We release the code, models and dataset at \url{https://github.com/jiesico/FusionRAFT}.
\end{abstract}


\section{Introduction}

\IEEEPARstart{O}{ptical} flow algorithms are essential for determining the motion of objects or regions within images between consecutive video frames. They generate a 2D vector field that describes the apparent movement of pixels over time. In contrast, scene flow focuses on estimating the pixel-level 3D motion in stereo or RGBD video frames \cite{Raft-3D}. These algorithms find wide applications in robotics \cite{FusionSLAM2020, navigation} and surveillance \cite{action_recognition1, action_recognition2}.
Computing optical flow becomes particularly challenging in environments with non-informative textures or when scenes are captured under low-lighting conditions. To address these difficulties, deep learning methods have emerged as effective solutions for optical flow estimation, formulating the problem as an energy minimization task \cite{Flow_Net,Flow_Net2.0,Flow_fields,Flow_fields++}. Deep learning-based optical flow approaches have demonstrated significant improvements over traditional methods \cite{Raft,CRaft,CamLiFlow}.

Several approaches utilize the computation of a correlation volume in the visible spectrum (RGB) to estimate the optical flow between two frames \cite{Flow_Net, Raft, CRaft}. The correlation volume captures inter-frame similarity by taking the dot product of the corresponding convolutional feature vectors and can be generated through an end-to-end deep network. This deep network can be designed to minimize an underlying energy function.
However, relying solely on RGB information can be limited in scenes affected by motion blurs, non-informative textures, or low illumination conditions. 
To address this limitation, some approaches have incorporated multimodal information. For example, depth or point cloud data can provide an alternative representation of the underlying scene structure. 
This multimodal information can be integrated through \textit{late fusion}, where feature vectors are combined without intermediate information exchange~\cite{DeepLiDARFlow, Raft-3D}, or through exchanging information between branches while sacrificing the independence of the single-modality representation~\cite{CamLiFlow}.

In this paper, we present a novel multimodal fusion approach, named  {\mname}, for optical and scene flow estimation, specifically designed to handle data captured in noisy or low-lighting conditions, for example those that can be encountered in search and rescue applications~\cite{Murphy2009}. 
Our approach introduces three key components to address these challenges.
Firstly, we propose a feature-level fusion technique that seamlessly blends RGB and depth information using a shared loss function.
Secondly, we introduce a self-attention mechanism that enhances the expressiveness of feature vectors by dynamically balancing the importance of features within each individual modality.
Lastly, we incorporate an optimized cross-attention module that facilitates information exchange and balance between RGB and depth modalities.
We integrate these new modules within RAFT~\cite{Raft} and RAFT-3D~\cite{Raft-3D}, using an application-oriented data augmentation strategy to learn robust feature representations that make optical and scene flow estimation effective in complex environments.
We conduct extensive evaluations on standard optical and scene flow benchmarks, as well as on two new settings that we introduce to assess robustness against photometric noise and challenging illumination conditions. 
Our method achieves state-of-the-art performance on the synthetic dataset FlyingThings3D \cite{flythings} and demonstrates superior generalization capabilities on the real-world dataset KITTI \cite{kitti} without fine-tuning.

\section{Related work}\label{sec:related_work}

We provide a comprehensive analysis of the recent progress in optical flow estimation using deep learning, followed by an in-depth investigation into the integration of multimodal fusion techniques for improving flow estimation performance.

\noindent \textbf{Optical flow.}
FlowNet \cite{Flow_Net} pioneered the use of deep neural networks to estimate optical flow as a supervised learning task.
FlowNet learns features across scales and abstraction levels to determine pixel correspondences.
FlowNet inspired FlowNet2.0 \cite{Flow_Net2.0}, PWC-Net \cite{PWC-Net}, MaskFlowNet \cite{Maskflownet} and LiteFlowNet3 \cite{LiteFlowNet3}.
FlowNet2.0 presents a warping operation and a method for stacking multiple networks through this operation \cite{Flow_Net2.0}.
PWC-Net utilizes pyramidal processing, warping, and a cost volume approach to improve both the size and accuracy of optical flow models \cite{PWC-Net}.
MaskFlowNet incorporates an asymmetric occlusion-aware feature matching module, which learns to filter out occluded regions through feature warping without the need for explicit supervision \cite{Maskflownet}.
LiteFlowNet3 tackles the challenge of estimating optical flow in the presence of partially occluded or homogeneous regions by using an adaptive affine transformation and a confidence map that identifies unreliable flow \cite{LiteFlowNet3}.
The confidence map is used to guide the generation of transformation parameters.

RAFT \cite{Raft} is a per-pixel feature extraction approach that constructs multi-scale 4D correlation volumes for each pixel pair, and updates the flow field iteratively through a recurrent unit.
Like FlowNet, RAFT has inspired GMA \cite{GMA} and CRAFT \cite{CRaft}. 
GMA addresses occlusions by modeling image self-similarities by using a global motion aggregation module, a transformer-based approach for finding long-range dependencies between pixels in the first image, and a global aggregation of the corresponding motion features. 
CRAFT aims to estimate the large motion displacements through a semantic smoothing transformer layer that integrates the features of one image and a cross-attention layer that replaces the original dot-product operator for correlation used in RAFT.
Unlike these approaches, we tackle the problem of estimating optical flow in situations of unreliable RGB information, such as noises and scarce illuminations, by appropriately fusing multiple modalities through self and cross attention within feature extraction layers.

\noindent \textbf{Multimodal fusion.}
Multimodal fusion can be performed at various stages: early-, mid-, and late-fusion. 
In early-fusion, multiple channels are created within the network to process multiple modalities together \cite{early-fusion}. 
Mid-fusion maintains different branches for each modality and then merges the corresponding features at the end of the network~\cite{mid-fusion1, mid-fusion2}. 
In late-fusion, the network is trained on each modality separately and then fuses the results from the independent branches~\cite{late-fusion}.
RAFT \cite{Raft}, GMA \cite{GMA}, and CRAFT \cite{CRaft} estimate the relationships between two consecutive frames using RGB images.
Inspired by multimodal fusion, some of these works have been improved to compute both scene and optical flow by utilizing additional modalities such as depth, and point clouds.

\subsubsection{RGB + Point Cloud Data.}
DeepLiDARFlow \cite{DeepLiDARFlow} exhibits improved performance in challenging conditions, such as reflective surfaces, poor illumination, and shadows.
Images and point clouds are processed by using multi-scale feature pyramid networks.
Late-fusion based on differentiable confidence volumes produces the fused features.
CamLiFlow \cite{CamLiFlow} improves upon DeepLiDARFlow by fusing dense image features and sparse point features more effectively.
Instead of late-fusion, CamLiFlow adopts a multi-stage, bidirectional fusion strategy, in which the two modalities are learned in separate branches using modality-specific architectures. CamLiRAFT \cite{camliraft2023} further improves the performance based on the RAFT \cite{Raft} framework, leading to superior results compared to CamLiFlow \cite{CamLiFlow}. 
Our method differs from previous methods in that it ensures the independence of each modality through the use of two separate branches and balances the information between the modalities through multi-stage information exchange.

\subsubsection{RGB + depth}
RAFT-3D \cite{Raft-3D} extends RAFT to estimate both optical and scene flow from RGBD data.
RGB images serve as inputs to the feature network, where a 4D correlation volume is constructed and a soft grouping of pixels into rigid objects is formed with the aid of depth information. 
Unlike RAFT \cite{Raft}, RAFT-3D employs late-fusion with the depth information and the RGB features in the prediction module, improving the stability of flow prediction. 
However, RAFT's feature extraction method may not sufficiently capture the rich 3D structural information. 
To address this, our approach employs early-fusion, in which features are extracted from both RGB and depth information, enabling stable estimation even in cases where RGB information is unreliable.

\section{Our approach}\label{sec:method}
\begin{figure*}[!t]
\centering
\small
\includegraphics[width=1.0\textwidth]{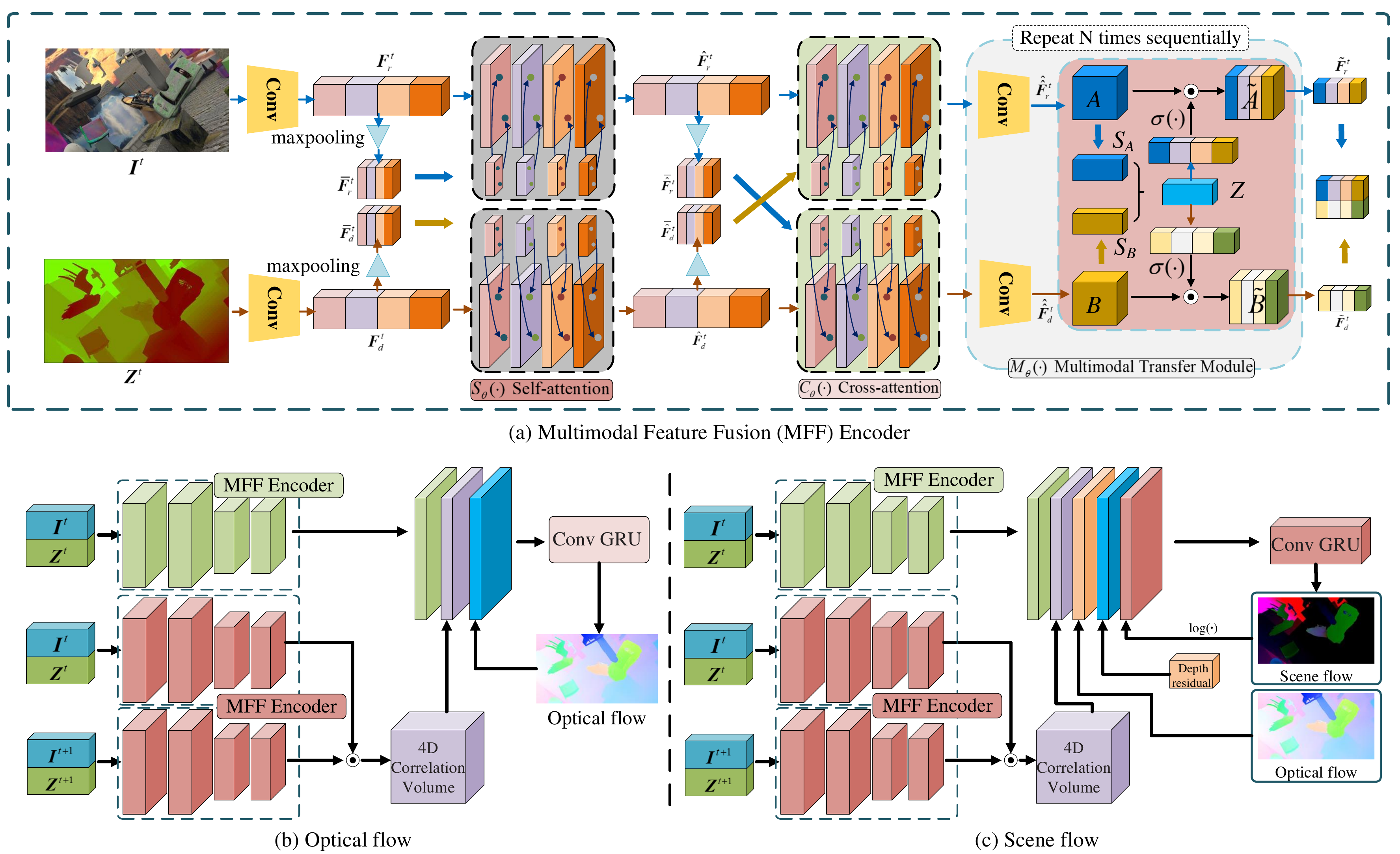}

\caption{Block diagram of \mname.
(a) Our encoder architecture: RGB and depth frames are taken as inputs.
The encoder network is a two-branch network with a transformer (self-attention plus cross-attention) and a Multimodal Transfer Module.
(b) Optical flow and (c) scene flow architectures. 
Two consecutive RGBD frames are taken as inputs by the MFF for the feature encoder, and the first RGBD frame is taken as input by the MFF for the context encoder.}
\label{fig:method:mmf}
\end{figure*}

We present a Multimodal Feature Fusion (MFF) Encoder that performs early fusion of RGB and depth modalities to improve the estimation of both optical and scene flow under noisy or poor lighting conditions.
Our encoder is flexible and can be integrated into flow estimation frameworks by replacing their original feature encoder. 
To achieve this, we employ self-attention, cross-attention, and Multimodal Transfer Module (MMTM) \cite{MMTM}. 
We extract low-level features from each modality and improve their expressivity using self-attention.
Cross-attention enables the network to attend to the most informative modality. 
MMTM is used to further fuse the attended features that are computed from the two modalities.
Fig.~\ref{fig:method:mmf}(a) shows the architecture of our encoder.

\subsection{Multimodal Feature Fusion Encoder} \label{sec:methods:fusion_encoder}

The Multimodal Feature Fusion Encoder takes a pair of consecutive RGBD frames $(P^t,~P^{t+1})$ at time $t$ as input. Each frame $P^t = \{I^t, Z^t\}$ is composed of a RGB image $I^t$ and a depth image $Z^t$.

We first obtain low-level features $\bm{{F}}^t_r\in \mathbb{R}^{W \times H \times D}$ and $\bm{{F}}^t_d\in \mathbb{R}^{W \times H \times D}$ from each modality with convolutional blocks, where we use the subscript ${r}$ to represent the RGB branch and ${d}$ for the depth branch (Fig.~\ref{fig:method:mmf}(a)).
${D}$ is the feature dimension and ${W \times H}$ is the resolution of the features.

\noindent\textbf{Self-attention.}
The local features $\bm{{F}}^t_r$ and $\bm{{F}}^t_d$ are obtained with convolutions that have limited receptive fields, therefore we model global structures by establishing long-range dependencies through a self-attention module (${S_\theta }(\cdot)$ in Fig.~\ref{fig:method:mmf}(a)).
To mitigate the high computational cost of self-attention, we downsample $\bm{{F}}^t_r \in \mathbb{R}^{N\times D}$ and $\bm{{F}}^t_d \in \mathbb{R}^{N\times D}$ to obtain $\bm{\bar{F}}^t_r$ and $\bm{\bar{F}}^t_d$ via $3\times3$ and $5\times5$ max-pooling layers.
With these downsampled features, we can use a multi-attention layer with four parallel attention heads to process $\bm{F}^t_r$ and $\bm{\bar{F}}^t_r$ (or $\bm{F}^t_d$ and $\bm{\bar{F}}^t_d$) in parallel and get $\bm{\hat{F}}^t_k$:

\vspace{-0.2cm}
\begin{equation}
\small
\begin{aligned}
&\bm{\hat{F}}^t_k \leftarrow{S_\theta }(\bm{{F}}^t_k, \bm{\bar{F}}^t_k)\\
& = \bm{{F}}^t_k{+}\mbox{MLP}\left(\sigma\left({\bm{W}^t_{Q_s}\bm{F}^t_k\left(\bm{W}^t_{K_s}\bm{\bar{F}}^t_k\right)^\top}\big/{\sqrt{D}}\right)\bm{W}^t_{V_s}\bm{\bar{F}}^t_k\right), 
\end{aligned}
\end{equation}
%
where $k\in\{r,d\}$ and $\sigma$ is the \textit{softmax} function. $D$ is the feature dimension. 
$\bm{W}^t_{Q_s}\in \mathbb{R}^{N\times D}, \bm{W}^t_{K_s}\in \mathbb{R}^{J\times D}$ and $\bm{W}^t_{V_s}\in \mathbb{R}^{J\times D}$ are the query, key and value matrices, where $N = W \times H$, $J = (W \times H)/(3 \times 5)$.
$\mbox{MLP}(\cdot)$ denotes a three-layer fully connected network with instance normalization\cite{ulyanov2016instance} and ReLU~\cite{xu2015empirical} activation after the first two layers.

\noindent\textbf{Cross-attention.}
We promote information exchange between the two modalities via cross-attention, which we implement through the network ${C_\theta }(\cdot)$ (Fig.~\ref{fig:method:mmf}(a)).
Attention signals from one modality (e.g.~RGB) emphasize the features of another modality (e.g.~depth), and vice versa. 
Given the self-attended features $\bm{\hat{F}}^t_r \in \mathbb{R}^{N\times D}$ and $\bm{\hat{F}}^t_d \in \mathbb{R}^{N\times D}$, we also adopt two downsampled networks max-pooling ($3\times3$), and max-pooling ($5\times5$) to generate the downsampled image feature map $\bm{\bar {\hat{F}}}^t_r$ (or $\bm{\bar {\hat{F}}}^t_d$). 
We denote the transformed features as $\bm{\hat{\hat{F}}}^t_r \in \mathbb{R}^{N\times D}$ and $\bm{\hat{\hat{F}}}^t_d \in \mathbb{R}^{N\times D}$ attained by cross-attention via
\begin{equation}
\small
\begin{aligned}
&\bm{\hat{\hat{F}}}^t_r \leftarrow{C_\theta }(\bm{\hat{F}}^t_r, \bm{\bar{\hat{F}}}^t_d)\\
& = \bm{\hat{F}}^t_r {+}\mbox{MLP}\left(\sigma\left(\bm{W}^t_{Q_c}\bm{\hat{F}}^t_r\left(\bm{W}^t_{K_c}\bm{\bar{\hat{F}}}^t_d\right)^\top\big/{\sqrt{D}}\right)\bm{W}^t_{V_c}\bm{\bar{\hat{F}}}^t_d\right), 
\end{aligned}
\end{equation}
where $W^t_{Q_c}\in \mathbb{R}^{N\times D}, W^t_{K_c}\in \mathbb{R}^{J\times D}$ and $W^t_{V_c}\in \mathbb{R}^{J\times D}$ are the query, key and value matrices.
This cross-attention block is also applied in the reverse direction so that information flows in both directions, i.e., RGB$\rightarrow$depth and depth$\rightarrow$RGB.

\noindent\textbf{Multimodal Transfer Module.}
Because our architecture operates with multimodal information, we further promote information exchange between modalities after attention.
Let ${M_\theta }(\cdot)$ be the Multimodal Transfer Module \cite{MMTM} we use to improve the balance between RGB and depth information (Fig.~\ref{fig:method:mmf}(a)).
Let $\bm{\hat{\hat{F}}}^t_r \in \mathbb{R}^{N\times D_M}$ and $\bm{\hat{\hat{F}}}^t_d \in \mathbb{R}^{N\times D_M}$ be the input multimodal features to MMTM, and $\bm{\tilde {F}}_r^t \in \mathbb{R}^{N\times D_M}$ and $\bm{\tilde {F}}_d^t \in \mathbb{R}^{N\times D_M}$ be the respective outputs.
MMTM first squeezes the feature vectors into ${S_{\bm{\tilde {F}}^t_r}}$ and ${S_{\bm{\tilde {F}}^t_d}}$ via a global average pooling.
MMTM then maps these tensors to a joint representation \emph{Z} through concatenation and a fully-connected layer.
Based on \emph{Z}, MMTM finally balances RGB and depth information by gating the channel-wise features:
\begin{equation}
\begin{aligned}
{S_{\bm{\tilde {F}}^t_{k}}} &= \frac{1}{{\Pi _{t = 1}^K{N_k}}}\sum\limits_{{n_{1, {\cdots} ,}}{n_K}} {\bm{\hat{\hat{F}}}^t_{k}({n_1}, \cdots ,{n_K})}, \\
{{Z}} &= \bm{\mbox{W}}[{{S_{\bm{\tilde {F}}^t_{r}}}},{{S_{\bm{\tilde {F}}^t_{d}}}}] + b \\
\tilde{\bm{ {F}}}^t_{k} &= 2\sigma ({\bm{\mbox{W}}_{\bm{\tilde {F}}^t_{k}}}{Z}) \odot \bm{\hat{\hat{F}}}^t_k,
\end{aligned}
\end{equation}
where $\left[\cdot,\cdot\right]$ is the concatenation operator and $k\in\{r,d\}$. $N_k$ represents the spatial dimensions of $\bm{\hat{\hat{F}}}^t_k$ and $D_M$ represents the number of channels of the features. 
$\bm{\mbox{W}}\in \mathbb{R}^{D_Z\times 2D_M}$, ${\bm{\mbox{W}}_{\bm{\tilde {F}}^i_{k}}}\in \mathbb{R}^{D_M\times D_Z}$ are the weights,
and $b \in \mathbb{R}^{D_Z}$ are the biases of the fully connected layers.

\subsection{Optical and scene flow estimation}
The inputs of optical and scene flow estimation are the feature vectors $[\tilde {\bm{{F}}}^t_{r},\tilde{\bm{ {F}}}^t_{d}]$ and $[\tilde {\bm{{F}}}^{t+1}_{r},\tilde{\bm{ {F}}}^{t+1}_{d}]$.
By calculating the dot product of feature vectors between the inputs, a 4D correlation volume \textbf{C} is generated:
\begin{equation}
\small
\begin{aligned}
    &fnet({P}^t) = [\tilde {\bm{{F}}}^t_{r},\tilde{\bm{ {F}}}^t_{d}] = [{M_\theta}({C_\theta}({S_\theta}({I^t}), {S_\theta}({Z^t})))],\\
    &\textbf{C}({P^t},{P^{t + 1}}) = \langle fnet({P^t}),fnet({P^{t + 1}})\rangle.
\end{aligned}
\end{equation}
where $\left<\cdot,\cdot\right>$ is the dot product operator. A four-layer pyramid $\{ {\textbf{C}_1},{\textbf{C}_2},{\textbf{C}_3},{\textbf{C}_4}\}$ is generated by reducing the last two dimensions of the correlation volume through pooling with kernels of size 1, 2, 4, and 8.

We compute 4D correlation volumes to estimate optical and scene flow \cite{Raft,Raft-3D}.
Through $\{{\textbf{C}_1},{\textbf{C}_2},{\textbf{C}_3},{\textbf{C}_4}\}$, 
we iteratively estimate the dense displacement field $\{\textbf{f}^1_{est},\textbf{f}^2_{est},...,\textbf{f}^M_{est}\}$ with M iterations to update the optical and scene flow.
We train our network by computing the loss between the estimated flow and the ground-truth flow ${\textbf{f}_{gt}}$ as
\begin{equation}
\small
\mathcal{L} = \sum\limits_{k = 1}^M {{\gamma ^{M - k}}{{\left\| {\textbf{f}^k_{est} - {\textbf{f}_{gt}}} \right\|}_1}},\
\label{eq:loss}
\end{equation}
where as the iteration $k$ increases, the weight per loss term exponentially increases with a base $\gamma$. 
Fig.~\ref{fig:method:mmf}(b,c) show how our Multimodal Feature Fusion Encoder is integrated in RAFT and RAFT-3D to estimate the optical flow and the scene flow, respectively.
Our module can be integrated seamlessly and does not require any modification to RAFT and RAFT-3D's modules after the 4D correlation volume computation.

\section{Experiments}

We compare {\mname} against state-of-the-art approaches on the FlyingThings3D~\cite{flythings} and KITTI~\cite{kitti} datasets. 
We design two experimental settings to mimic corrupted RGB images and poor lighting condition scenarios.
We also evaluate on data we acquired with a RGBD sensor in various lighting conditions.
We report both quantitative and qualitative results, and carry out ablation studies.

\subsection{Experimental setup}

\noindent \textbf{Datasets.}
FlyingThings3D~\cite{flythings} is split into \textit{clean} and \textit{final} sets containing dynamic synthetic scenes.
The former is composed of 27K RGBD images including changing lighting and shading effects, while the latter is an augmented version of the former with simulated challenging motions and blurs.
Each set contains train and test splits. 
Previous methods~\cite{Raft,Raft-3D,CamLiFlow} exclude samples containing fast-moving objects during the evaluation. However, as such visual challenges is of interests to our problem, we use the \textit{whole} training set of FlyingThings3D and sample 1K RGBD image pairs from the \textit{whole} test set for the evaluation.
KITTI consists of real-world scenes captured from vehicles in urban scenarios.
Because the original dataset does not provide depth data, we use the disparity estimated by GA-Net~\cite{Ga-net} as in~\cite{Raft-3D}.
We exploit KITTI to assess the ability of our model and the compared ones in generalizing from synthetic to real data, without training or finetuning using any of the KITTI’s sequences. 
We use the training set of KITTI as our evaluation set since KITTI's test set is not publicly available.
To further validate the performance of {\mname} in real-world scenarios, we collect an RGBD dataset using a Realsense D415 camera in an indoor office with moving people under three lighting setups, named Bright, Dimmed, and Dark.
The Bright setting features bright lighting, where the moving objects are clearly visible.
The Dimmed setting features dimmed lighting, where the moving objects can be observed with a lower visual quality.
The Dark setting features very low lighting where the moving objects can be barely seen.
We only qualitatively evaluate this dataset because we could not produce optical flow ground truth.

\noindent \textbf{Evaluation metrics.}
We quantify the optical and scene flow results using conventional evaluation metrics \cite{Raft, CRaft, Raft-3D}:
for the optical flow we use $\rm AEPE_{2D}$(pixel), $\rm {ACC}_{1px}$(\%) and $\rm Fl^{all}_{2D}$(\%), 
for the scene flow we use $\rm AEPE_{3D}$(m), $\rm {ACC}_{0.05m}$(\%), $\rm {ACC}_{0.10m}$(\%) and $\rm Fl^{all}_{3D}$(\%).
$\rm AEPE_{2D}$ measures the average end-point error (EPE)~\cite{Raft}, which is an average value of all the 2D flow errors.
$\rm AEPE^{epe{\textless{100}}}_{2D}$ measures the average end-point error (EPE) among the 2D flow errors that are less than 100 pixels. 
$\rm AEPE_{3D}$ is the average of euclidean distance (EPE for 3D) between the ground-truth 3D scene flow and the predicted results.
$\rm AEPE^{epe{\textless{1}}}_{3D}$ measures the average end-point error (EPE) among the 3D flow errors that are less than 1 meter.
$\rm {ACC}_{1px}$ \cite{Raft-3D} measures the portion of errors that are within a threshold of one pixel.
$\rm {ACC}_{0.05m}$ \cite{Raft-3D} measures the portion of errors that are within a threshold of 0.05 meters, while $\rm {ACC}_{0.10m}$ \cite{Raft-3D} measures the portion of errors that are within a threshold of 0.10 meters.
$\rm MEAN_{AEPE}$ and $\rm MEAN_{ACC}$ are the average values of $\rm AEPE^{all}_{2D}$ and $\rm ACC_{1px}$, respectively, calculated over FlyingThings3D-clean and FlyingThings3D-final.
$\rm Fl^{all}_{2D}$ \cite{CRaft} is the percentage of outlier pixels whose end-point error is $>3$ pixels or $5\%$ of the ground-truth flow magnitude.
$\rm Fl^{all}_{3D}$~\cite{Flownet3d++} is the percentage of outlier pixels whose 3D Euclidean distance between the ground-truth 3D scene flow and the predicted one is $>0.3$ m or $5\%$ of the ground-truth flow magnitude.

\noindent \textbf{Evaluation settings.}
Environments with poor light conditions lead to weak texture information that can compromise the stability of feature representation.
Also additive Gaussian noises can affect optical and scene flow estimation.
To assess the robustness, we design three experimental settings on the public FlyingThings3D and KITTI datasets:
\emph{Standard}: we use the original version of the dataset;
\emph{AGN}: we apply Additive Gaussian Noise on RGB images;
\emph{Dark}: we darken RGB images.
In AGN we randomly sample noise values ($\alpha$) from a normal distribution centered in zero with a standard deviation equal to 35.
In Dark we divide pixel values by a random factor $\beta  \sim \mbox{U}(\{ 1,2, \cdots,9\} )$.

\noindent \textbf{Implementation details.}
We implemented \mname{} in PyTorch with all modules initialized with random weights.
We train our network for 100K iterations with the batch size of 6 on 3 Nvidia 3090 GPUs.
During training, we set the initial learning rate at $1.25\cdot10^{-4}$ and use linear decay. 
We apply MMTM sequentially with N {=} 3 times as suggested in the original paper~\cite{MMTM}.
We set $\gamma {=} 0.8$ in Eq.~\eqref{eq:loss} as in RAFT~\cite{Raft}.

\subsection{Comparisons}\label{sec:exp:comparisons}
We compare \mname{} against RGB methods for 2D optical flow estimation, i.e.~RAFT~\cite{Raft}, GMA~\cite{GMA}, CRAFT~\cite{CRaft}, and Separable flow~\cite{Separable_flow}, and against methods for 3D scene flow estimation, i.e.~RAFT-3D~\cite{Raft-3D} and CamLiRAFT~\cite{camliraft2023}.
See Sec.~\ref{sec:related_work} for the description of these methods.

\begin{table*}[t]
    \centering
    \caption{
    Optical flow estimation in the Standard, AGN, and Dark settings on FlyingThings3D-clean, FlyingThings3D-final, and KITTI. 
    All models are trained with FlyingThings3D, without fine-tuning on KITTI.
    Bold font indicates the best-performing method.}
    \label{tab:setting1_results}
    \vspace{-.2cm}
    \resizebox{\linewidth}{!}{%
    \begin{tabular}{clc|ccc|ccc|cc}
        \toprule
        & \multirow{2}{*}{Method} & \multirow{2}{*}{Input} & \multicolumn{3}{c|}{FlyingThings3D-clean}  &  \multicolumn{3}{c|}{FlyingThings3D-final}&  \multicolumn{2}{c}{KITTI-Train} \\
        & & & $\rm ACC_{1px}$&  $\rm AEPE^{epe{\textless{100}}}_{2D}$& $\rm AEPE^{all}_{2D}$ &$\rm ACC_{1px}$&  $\rm AEPE^{epe{\textless{100}}}_{2D}$& $\rm AEPE^{all}_{2D}$ &$\rm AEPE^{all}_{2D}$&$\rm Fl^{all}_{2D}$\\
        \midrule
        \multirow{8}{*}{\rotatebox[origin=c]{90}{Standard setting}} & \texttt{\footnotesize RAFT} \cite{Raft} &RGB & 77.06 &2.65 & 4.69 & 76.91&2.67 &4.39&6.76& 20.99 \\
        & \texttt{\footnotesize GMA} \cite{GMA} &RGB & 78.81& 2.58& 4.43 & 78.66& 2.57 &4.20& 6.10& 20.47 \\
        & \texttt{\footnotesize Separable flow} \cite{Separable_flow} & RGB & 75.39&2.88 & 4.57 &75.29 &2.84 &4.29&6.40& 20.66\\
        & \texttt{\footnotesize CRAFT} \cite{CRaft}& RGB & 77.90&2.79 & 4.85 & 77.70& 2.77& 4.66&6.82&21.95 \\
        & \texttt{\footnotesize {\mname}-2D} & RGBD & \textbf{80.37}&\textbf{2.17} & \textbf{3.52} & \textbf{80.21}&\textbf{2.22}  & \textbf{3.42}&\textbf{5.49}&\textbf{18.05}\\
        \cmidrule{2-11}
        & \texttt{\footnotesize RAFT-3D} \cite{Raft-3D}& RGBD & 86.01& 1.79& 3.58&85.97& 1.76& 3.57&5.91&17.80\\
        & \texttt{\footnotesize CamLiRAFT} \cite{camliraft2023}& RGB+LiDAR & 83.59& 1.81& 3.03&83.26& 1.80& 2.84& 4.84&14.76 \\
        & \texttt{\footnotesize {\mname}-3D} & RGBD & \textbf{87.45}&\textbf{1.57} & \textbf{2.58} & \textbf{87.39}&\textbf{1.58} & \textbf{2.69}&\textbf{4.70}&\textbf{12.36}\\
        \midrule
        \multirow{8}{*}{\rotatebox[origin=c]{90}{AGN setting}} & \texttt{\footnotesize RAFT} \cite{Raft} &RGB & 71.42&2.98 & 4.89 & 71.01&2.96 & 4.64& 7.23&23.45 \\
        & \texttt{\footnotesize GMA} \cite{GMA} &RGB & 72.63& 2.99& 5.23 & 72.20& 2.94& 5.24&7.01& 23.52 \\
        & \texttt{\footnotesize Separable flow} \cite{Separable_flow} & RGB & 68.95&3.15  & 5.32 &68.57 &3.18 & 5.23&8.26&25.79 \\
        & \texttt{\footnotesize CRAFT} \cite{CRaft} & RGB & 73.30&2.86 & 4.65 & 72.81& 2.88& 4.67& 7.45&23.65 \\
        & \texttt{\footnotesize {\mname}-2D} & RGBD & \textbf{77.24}&\textbf{2.24} &\textbf{3.50}  & \textbf{76.77} &\textbf{2.30}  & \textbf{3.38}&\textbf{5.47}&\textbf{19.15}\\
        \cmidrule{2-11}
        & \texttt{\footnotesize RAFT-3D} \cite{Raft-3D} & RGBD & 84.59 & 1.79 & 3.14&84.26  & 1.84 &3.21&5.50&17.94\\
        & \texttt{\footnotesize CamLiRAFT} \cite{camliraft2023} & RGB+LiDAR & 76.98 & 2.23 & 3.98&76.31 & 2.33 &3.71&5.26&16.98\\
        & \texttt{\footnotesize {\mname}-3D} & RGBD &\textbf{86.75} &\textbf{1.55}&\textbf{2.63} &\textbf{86.71} & \textbf{1.53}  &\textbf{2.60}&\textbf{4.53}&\textbf{11.57} \\
        \midrule
        \multirow{8}{*}{\rotatebox[origin=c]{90}{Dark setting}} & \texttt{\footnotesize RAFT} \cite{Raft} &RGB & 60.26&4.00 & 8.15 & 60.36& 4.01& 7.85& 11.75&31.81 \\
        & \texttt{\footnotesize GMA} \cite{GMA} &RGB &63.36&4.50 & 9.96 & 62.10&4.62 & 10.34&9.65&27.87 \\
        & \texttt{\footnotesize Separable flow} \cite{Separable_flow}& RGB & 68.20& 4.84& 7.96 &68.03 &4.86 & 7.79&10.09&28.41 \\
        & \texttt{\footnotesize CRAFT} \cite{CRaft} & RGB & 70.07&4.84 & 8.46 & 69.77& 4.87&8.44& 11.10&29.47  \\
        & \texttt{\footnotesize {\mname}-2D} & RGBD & \textbf{76.70}&\textbf{2.39} & \textbf{3.65} & \textbf{76.57}& \textbf{2.38} &\textbf{3.66}&\textbf{8.29}&\textbf{23.87} \\
        \cmidrule{2-11}
        & \texttt{\footnotesize RAFT-3D} \cite{Raft-3D} & RGBD & 81.03& 2.20&3.78 & 80.96&2.20  &3.56&15.14&32.08 \\
        & \texttt{\footnotesize CamLiRAFT} \cite{camliraft2023} & RGB+LiDAR & 74.80& 2.54&4.54 &74.73&2.64  &4.11&7.44 & \textbf{16.97}\\
        & \texttt{\footnotesize {\mname}-3D} & RGBD & \textbf{87.11}&\textbf{1.55} & \textbf{2.91} & \textbf{87.03}& \textbf{1.58} &\textbf{2.84}&\textbf{7.26}&20.07 \\
        \bottomrule 
    \end{tabular}
    }
\end{table*}
\begin{table*}[!h]
    \centering
    \caption{
    Scene flow estimation in the Standard, AGN, and Dark settings on FlyingThings3D-clean, FlyingThings3D-final, and KITTI.
    All models are trained with FlyingThings3D, without fine-tuning on KITTI.
    Bold font indicates the best-performing method.}
    \vspace{-.2cm}
    \label{tab:4}  
    \resizebox{\linewidth}{!}{%
    \begin{tabular}{lc|cccc|cccc|cc}
        \toprule
        \multirow{2}{*}{Method} & \multirow{2}{*}{Setting} & \multicolumn{4}{c|}{FlyingThings3D-clean}  &  \multicolumn{4}{c|}{FlyingThings3D-final}&  \multicolumn{2}{c}{KITTI-Train}  \\
         & & $\rm ACC_{0.05m}$& $\rm ACC_{0.10m}$&  $\rm AEPE^{epe{\textless{1}}}_{3D}$& $\rm AEPE^{all}_{3D}$& $\rm ACC_{0.05m}$ &$\rm ACC_{0.10m}$& $\rm AEPE^{epe{\textless{1}}}_{3D}$& $\rm AEPE^{all}_{3D}$&$\rm AEPE^{all}_{3D}$&$\rm Fl^{all}_{3D}$\\
        \midrule
        \texttt{\footnotesize RAFT-3D} \cite{Raft-3D}& Standard & 74.01& 81.22&0.064 & 0.186&74.25  &81.43&0.064&0.180&0.136 &5.20 \\
        \texttt{\footnotesize CamLiRAFT} \cite{camliraft2023} & Standard & 76.83& \textbf{87.98}&\textbf{0.049} & 0.104&\textbf{76.87} &\textbf{88.20}&\textbf{0.049}&0.102& \textbf{0.121}& 7.13\\
        \texttt{\footnotesize {\mname}-3D} & Standard & \textbf{77.04}&83.74 & 0.056 & \textbf{0.100}& 76.80 &83.58&0.057&\textbf{0.101}&0.134&\textbf{4.90} \\
        \midrule
        \texttt{\footnotesize RAFT-3D}\cite{Raft-3D} & AGN & 75.05& 81.77&0.065 & 0.193&74.75  &81.56&0.066&0.144&0.134&5.36\\
        
        \texttt{\footnotesize CamLiRAFT} \cite{camliraft2023}& AGN & 73.83& \textbf{86.59}&\textbf{0.055} & 0.116&73.38 &\textbf{86.31}&\textbf{0.055}&0.126& \textbf{0.122}& 7.81\\
        
        \texttt{\footnotesize {\mname}-3D} & AGN & \textbf{76.60}&82.71 & 0.061 & \textbf{0.104}& \textbf{76.35} &82.55&0.062&\textbf{0.107}&0.134&\textbf{5.52} \\
        \midrule
        \texttt{\footnotesize RAFT-3D}\cite{Raft-3D} & Dark & 71.21& 79.33&0.071& 0.203&71.00  &79.24&0.072&0.145&0.145&9.39 \\
        \texttt{\footnotesize CamLiRAFT}\cite{camliraft2023} & Dark & 66.70& 82.24&0.068& \textbf{0.141}&66.65  &82.01&0.069&0.127&0.171 &9.39 \\
        \texttt{\footnotesize {\mname}-3D} & Dark & \textbf{76.72}&\textbf{83.62} & \textbf{0.057} & 0.175& \textbf{76.58} &\textbf{83.51}&\textbf{0.057}&\textbf{0.112}&\textbf{0.136}&\textbf{6.20} \\
        \bottomrule 
    \end{tabular}
    }
\end{table*}

\subsubsection{Quantitative results}

Tab.~\ref{tab:setting1_results} (top) reports optical flow results in Standard setting.
{\mname}-2D outperforms GMA by $+1.56\%$ and $+1.55\%$ in terms of $\rm ACC_{1px}$, and $+0.91$ and $+0.78$ in terms of $\rm AEPE^{all}_{2D}$ in FlyingThings3D-clean and FlyingThings3D-final, respectively.
{\mname}-3D outperforms RAFT-3D by $+1.44\%$ and $+1.42\%$ in terms of $\rm ACC_{1px}$, and $+1.00$ and $+0.88$ in terms of $\rm AEPE^{all}_{2D}$. 
While RAFT-3D extracts features only from RGB images, our MFF encoder extracts features from both RGB and depth, producing more informative internal representations.
{\mname}-3D outperforms CamLiRAFT by $+3.86\%$ and $+4.13\%$ in terms of $\rm ACC_{1px}$, and $+0.45$ and $+0.15$ in terms of $\rm AEPE^{all}_{2D}$.

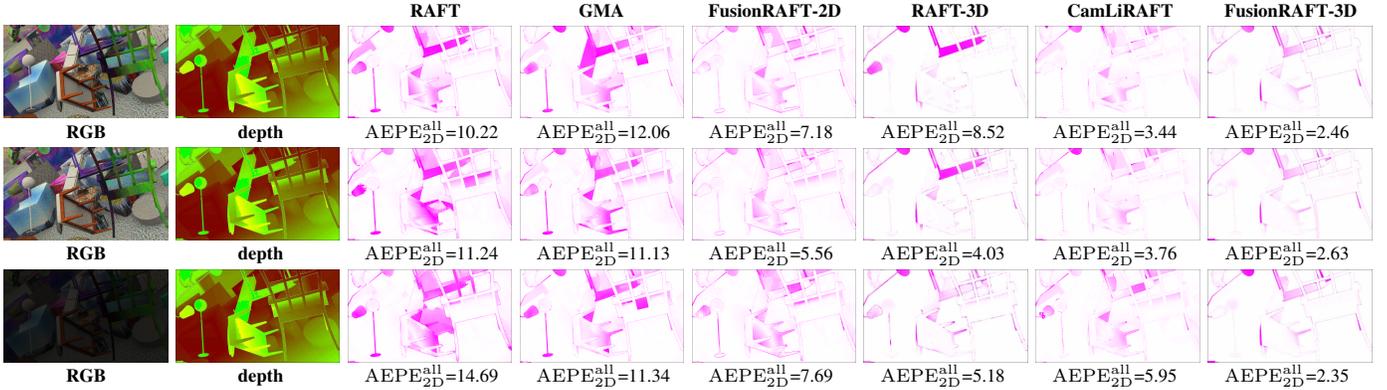
\begin{figure*}[t]
\vspace*{3mm}
\small
\begin{center}
  \begin{tabular}{@{}c@{}c@{}c@{}c@{}c@{}c}
    \vspace*{3mm}
    \begin{overpic}[width=.12\linewidth]{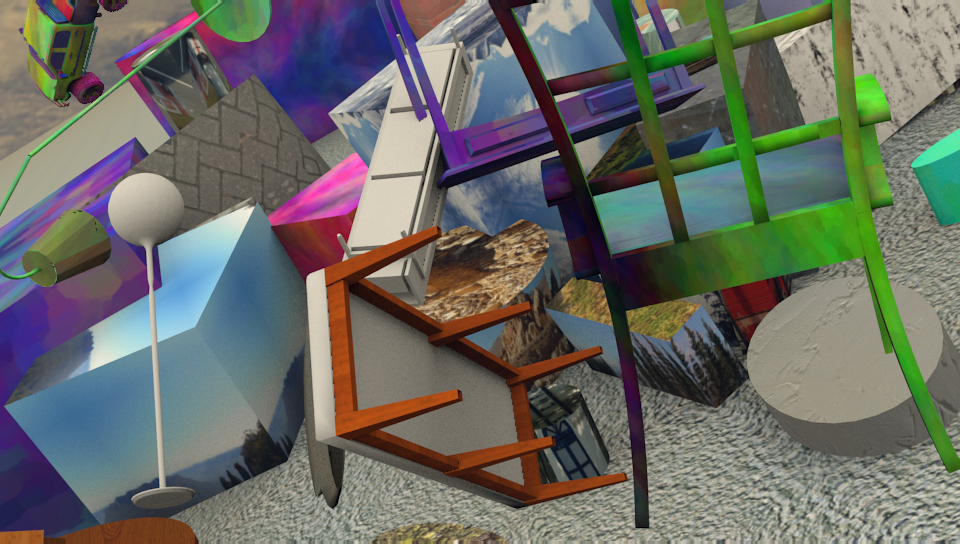}
     \put(38,-12){\color{black}\scriptsize\textbf{RGB}}
    \end{overpic}
    
    \begin{overpic}[width=.12\linewidth]{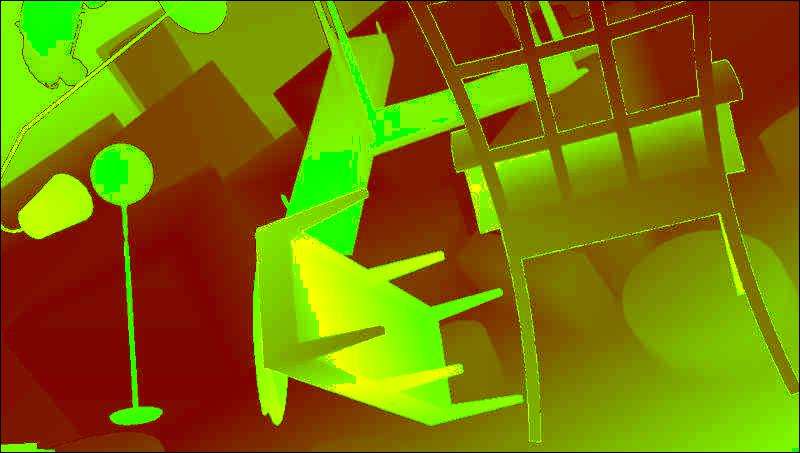}
    \put(38,-12){\color{black}\scriptsize\textbf{depth}}
    \end{overpic}
    \begin{overpic}[width=.12\linewidth]{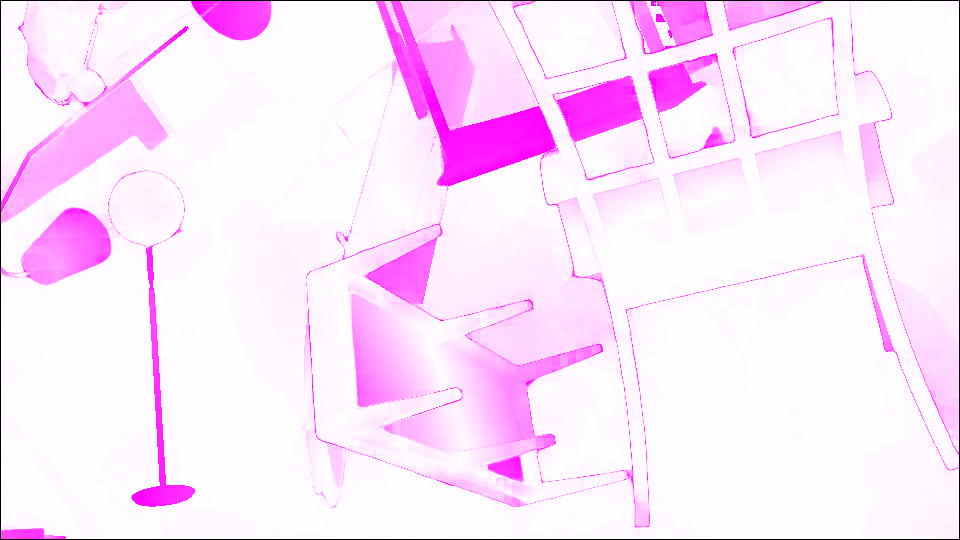}
    \put(38,62){\color{black}\scriptsize\textbf{RAFT}}
    \put(10,-12){\color{black}\scriptsize$\rm AEPE^{all}_{2D}$=10.22}
    \end{overpic}
    \begin{overpic}[width=.12\linewidth]{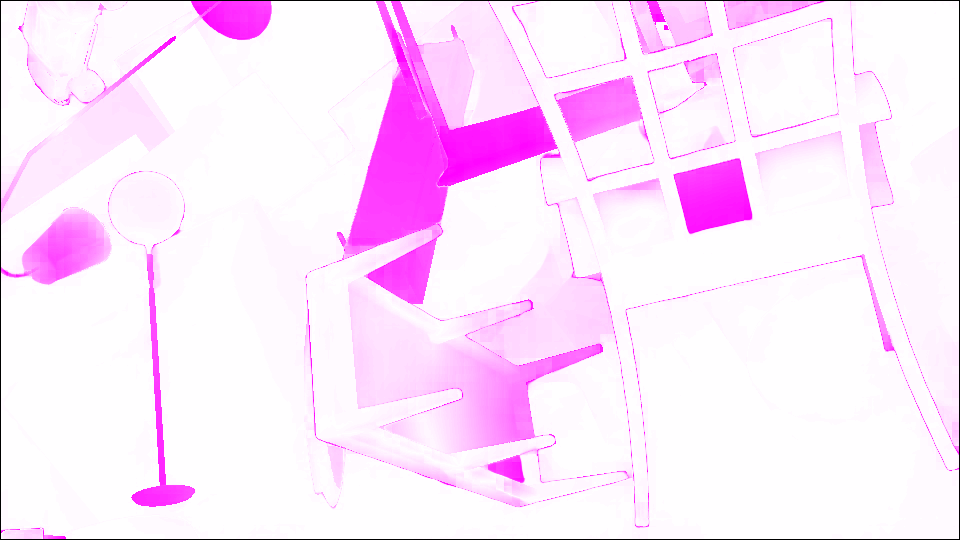}
    \put(36,62){\color{black}\scriptsize\textbf{GMA}}
    \put(10,-12){\color{black}\scriptsize$\rm AEPE^{all}_{2D}$=12.06}
     \end{overpic}
    \begin{overpic}[width=.12\linewidth]{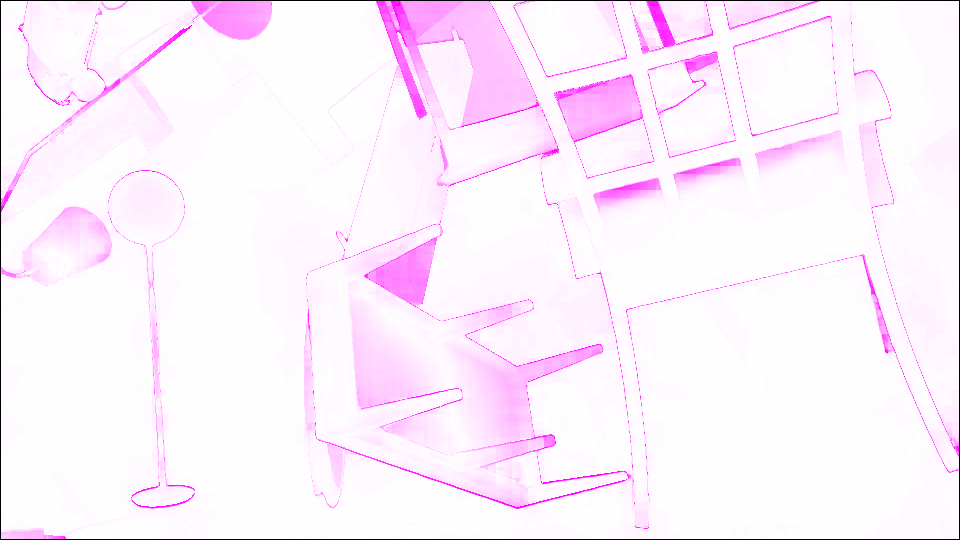}
    \put(10,62){\color{black}\scriptsize\textbf{{\mname}-2D}}
    \put(10,-12){\color{black}\scriptsize$\rm AEPE^{all}_{2D}$=7.18}
    \end{overpic}
    \begin{overpic}[width=.12\linewidth]{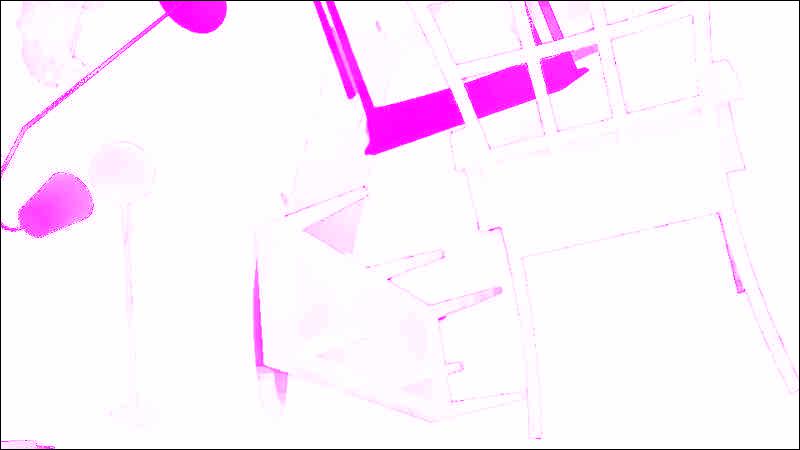}
    \put(29,62){\color{black}\scriptsize\textbf{RAFT-3D}}
    \put(10,-12){\color{black}\scriptsize$\rm AEPE^{all}_{2D}$=8.52}
    \end{overpic}
    \begin{overpic}[width=.12\linewidth]{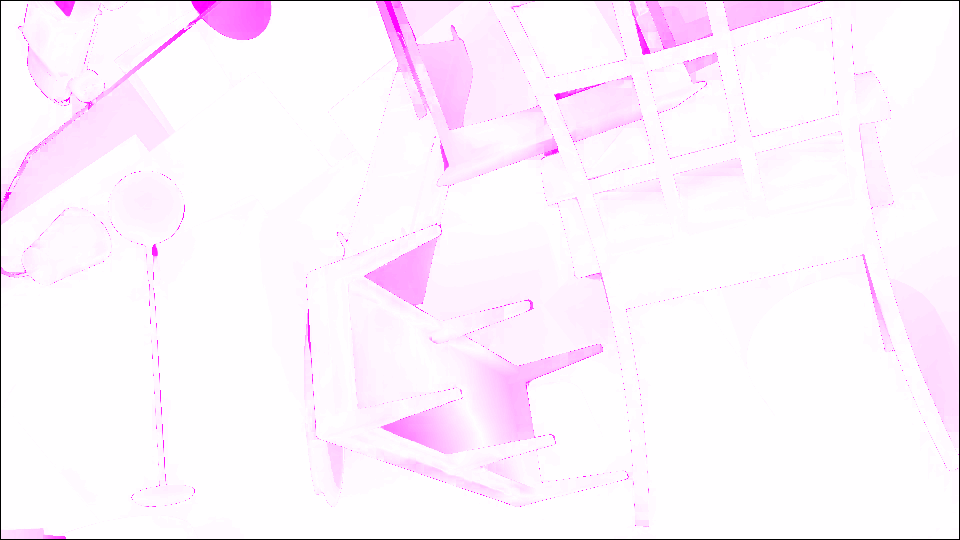}
    \put(19,62){\color{black}\scriptsize\textbf{CamLiRAFT}}
    \put(10,-12){\color{black}\scriptsize$\rm AEPE^{all}_{2D}$=3.44}
    \end{overpic}
    \begin{overpic}[width=.12\linewidth]{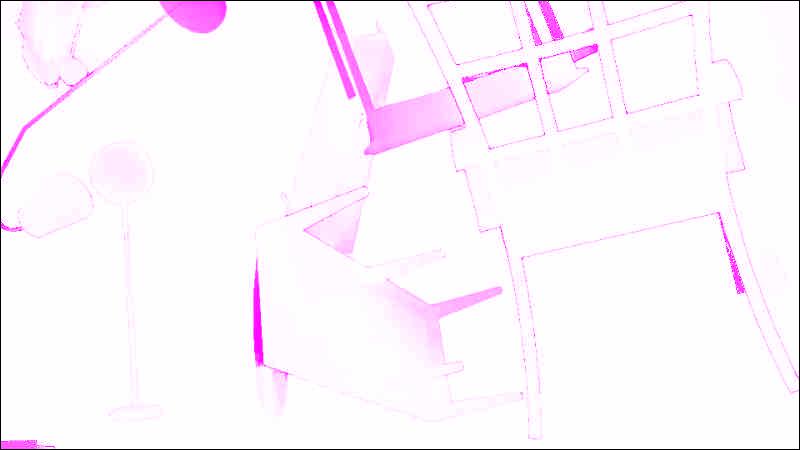}
    \put(10,62){\color{black}\scriptsize\textbf{{\mname}-3D}}
    \put(10,-12){\color{black}\scriptsize$\rm AEPE^{all}_{2D}$=2.46}
    \end{overpic}\\
    \vspace*{3mm}
    \begin{overpic}[width=.12\linewidth]{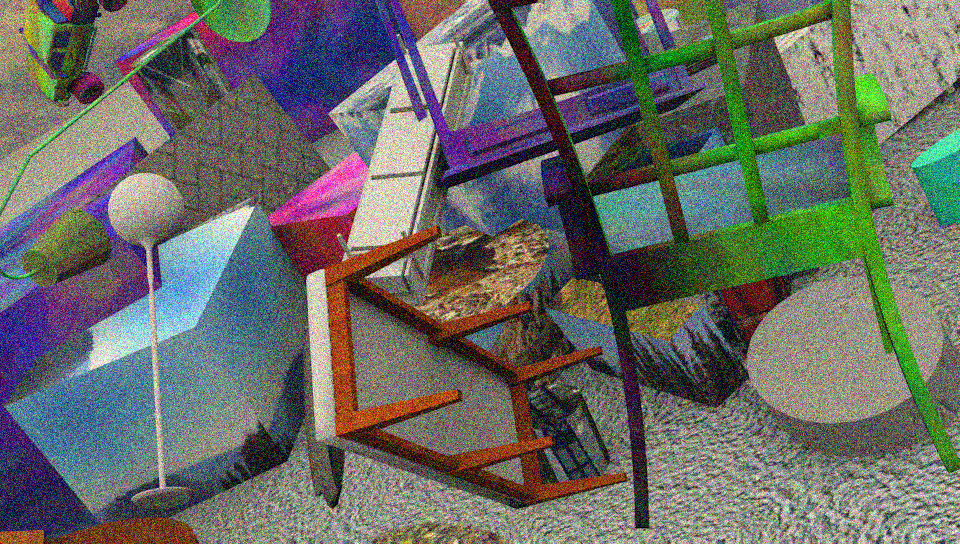}
    \put(38,-12){\color{black}\scriptsize\textbf{RGB}}
    \end{overpic}
    \begin{overpic}[width=.12\linewidth]{sections/images/flow-error/depth.jpg}
    \put(38,-12){\color{black}\scriptsize\textbf{depth}}
    \end{overpic}
    \begin{overpic}[width=.12\linewidth]{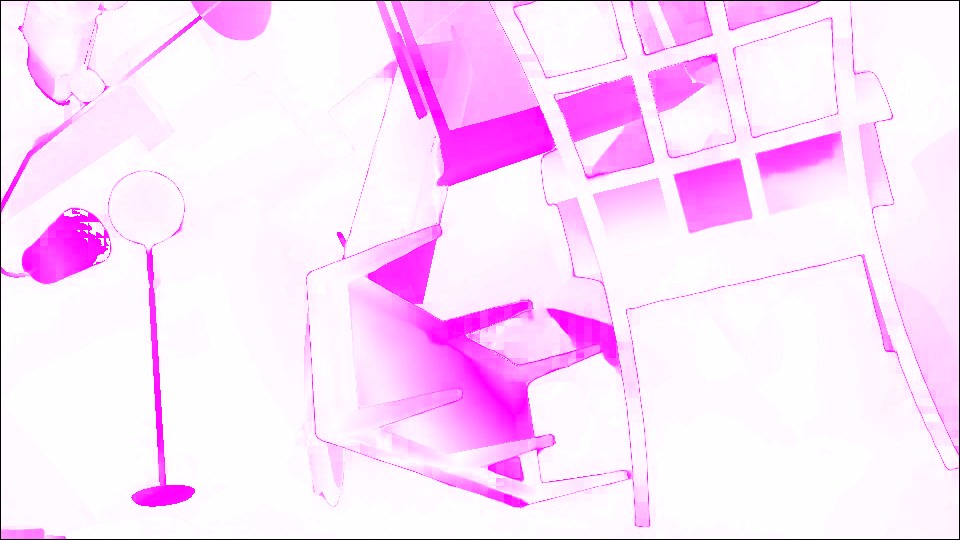}
    \put(10,-12){\color{black}\scriptsize$\rm AEPE^{all}_{2D}$=11.24}
    \end{overpic}
    \begin{overpic}[width=.12\linewidth]{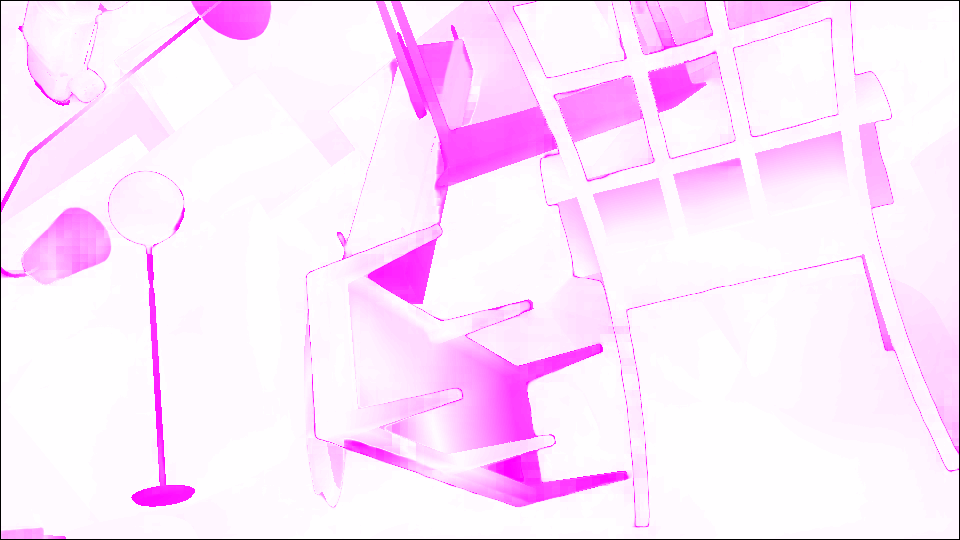}
    \put(10,-12){\color{black}\scriptsize$\rm AEPE^{all}_{2D}$=11.13}
     \end{overpic}
    \begin{overpic}[width=.12\linewidth]{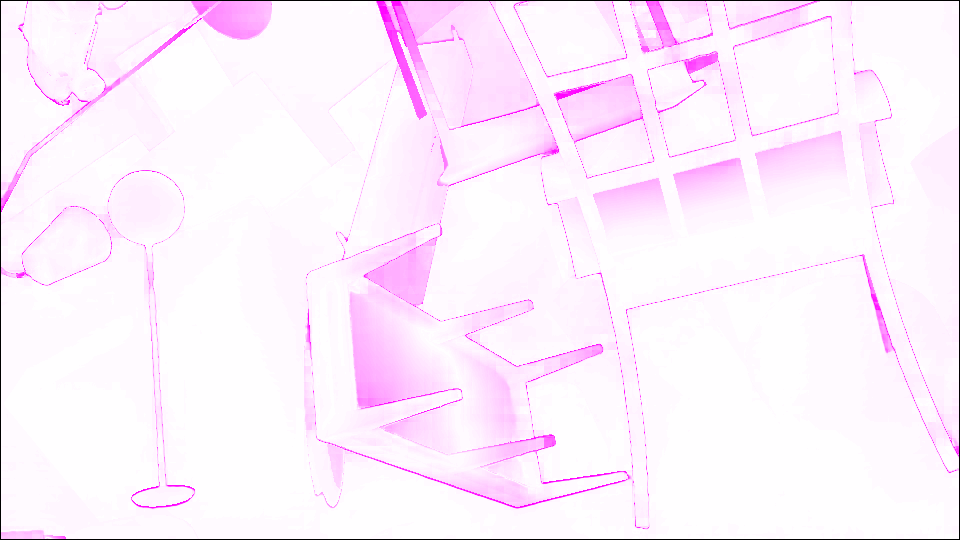}
    \put(10,-12){\color{black}\scriptsize$\rm AEPE^{all}_{2D}$=5.56}
    \end{overpic}
    \begin{overpic}[width=.12\linewidth]{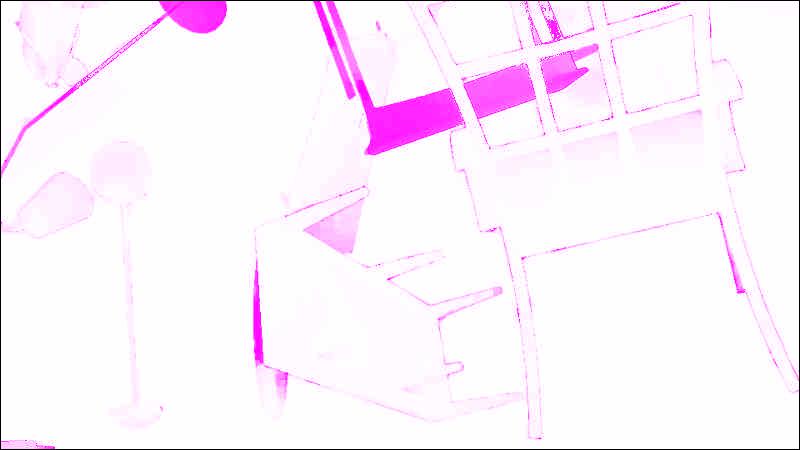}
    \put(10,-12){\color{black}\scriptsize$\rm AEPE^{all}_{2D}$=4.03}
    \end{overpic}
    \begin{overpic}[width=.12\linewidth]{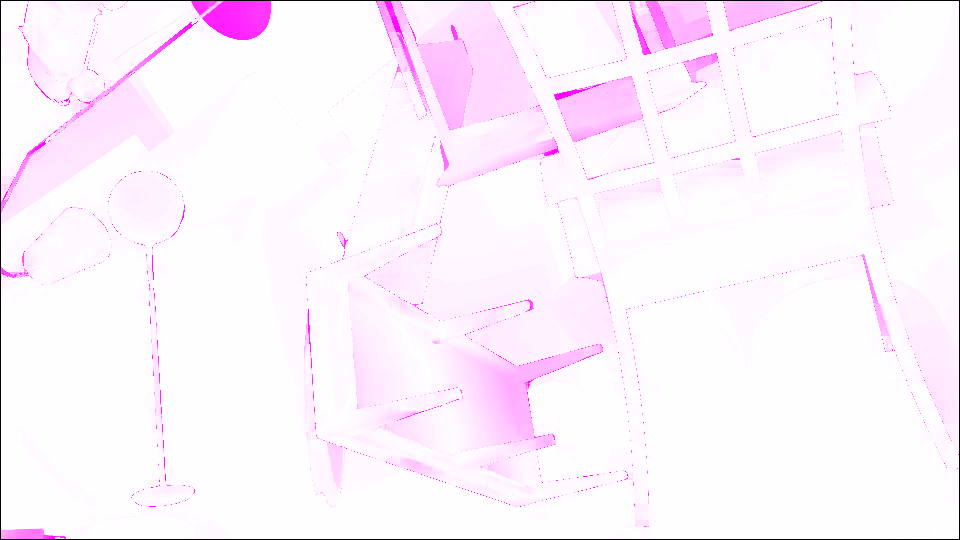}
    \put(10,-12){\color{black}\scriptsize$\rm AEPE^{all}_{2D}$=3.76}
    \end{overpic}
    \begin{overpic}[width=.12\linewidth]{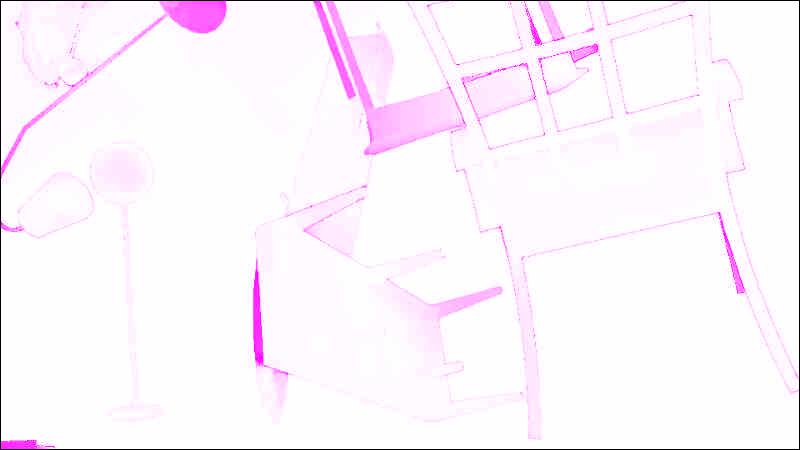}
    \put(10,-12){\color{black}\scriptsize$\rm AEPE^{all}_{2D}$=2.63}
    \end{overpic}\\
    \begin{overpic}[width=.12\linewidth]{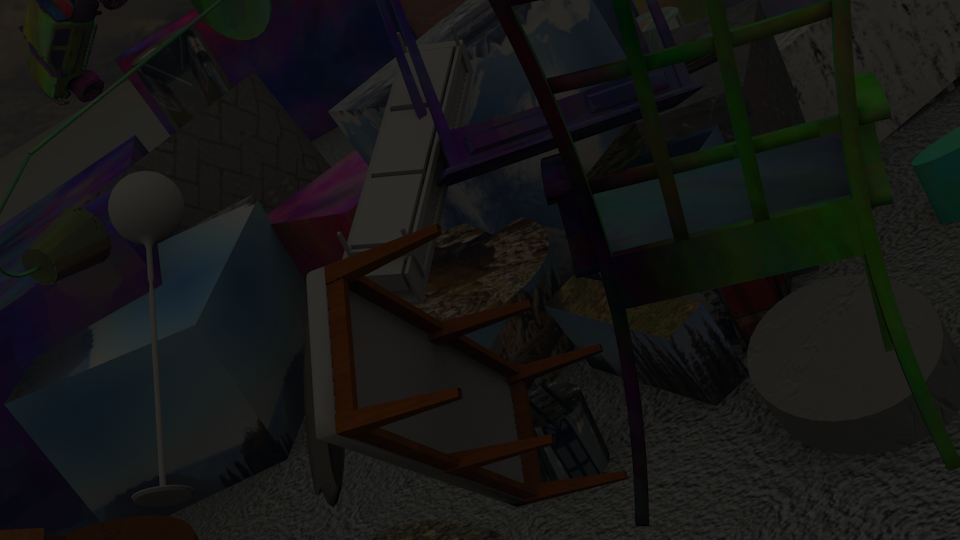}
    \put(38,-12){\color{black}\scriptsize\textbf{RGB}}
    \end{overpic}
    \begin{overpic}[width=.12\linewidth]{sections/images/flow-error/depth.jpg}
    \put(38,-12){\color{black}\scriptsize\textbf{depth}}
    \end{overpic}
    \begin{overpic}[width=.12\linewidth]{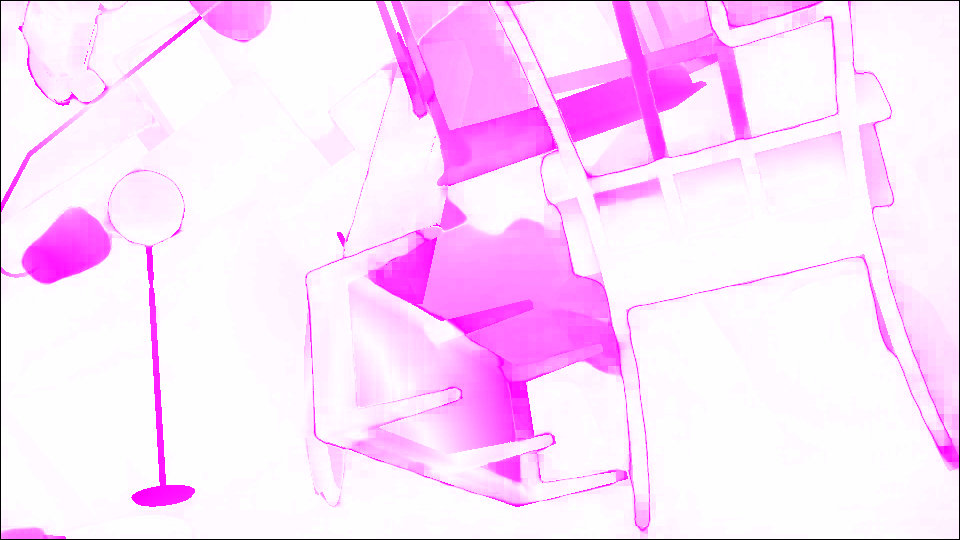}
    \put(10,-12){\color{black}\scriptsize$\rm AEPE^{all}_{2D}$=14.69}
    \end{overpic}
    \begin{overpic}[width=.12\linewidth]{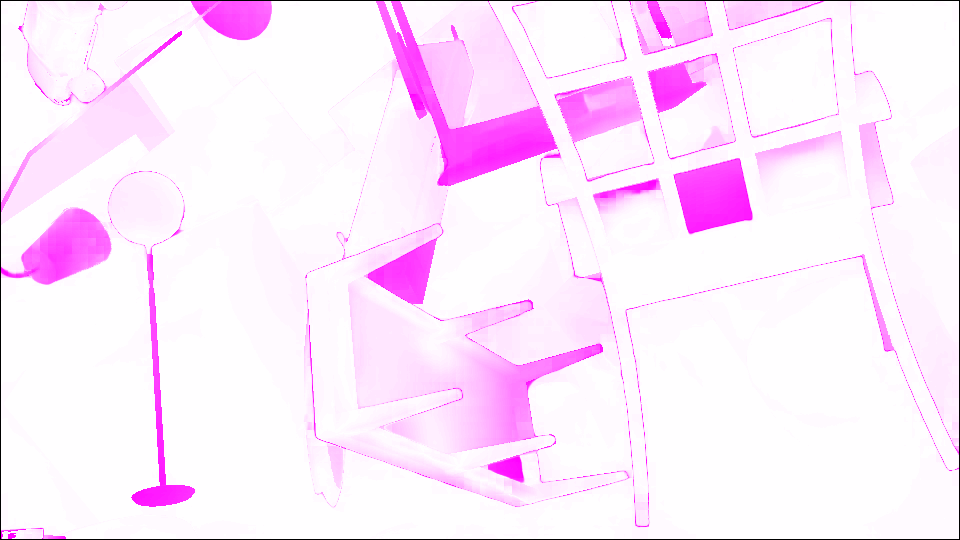}
    \put(10,-12){\color{black}\scriptsize$\rm AEPE^{all}_{2D}$=11.34}
     \end{overpic}
    \begin{overpic}[width=.12\linewidth]{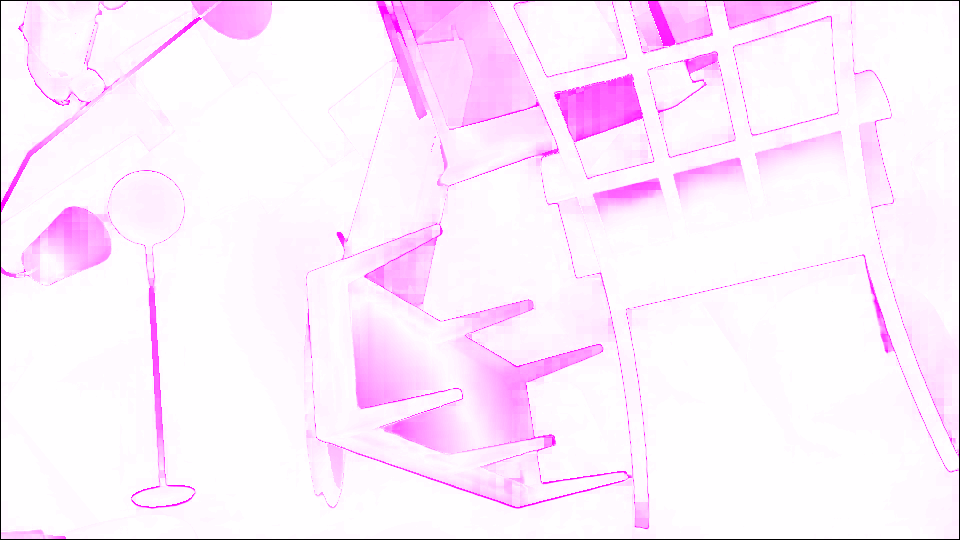}
    \put(10,-12){\color{black}\scriptsize$\rm AEPE^{all}_{2D}$=7.69}
    \end{overpic}
    \begin{overpic}[width=.12\linewidth]{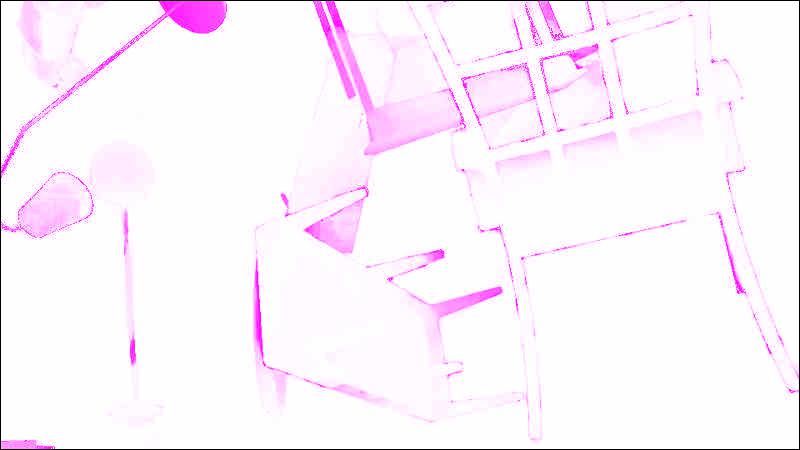}
    \put(10,-12){\color{black}\scriptsize$\rm AEPE^{all}_{2D}$=5.18}
    \end{overpic}
    \begin{overpic}[width=.12\linewidth]{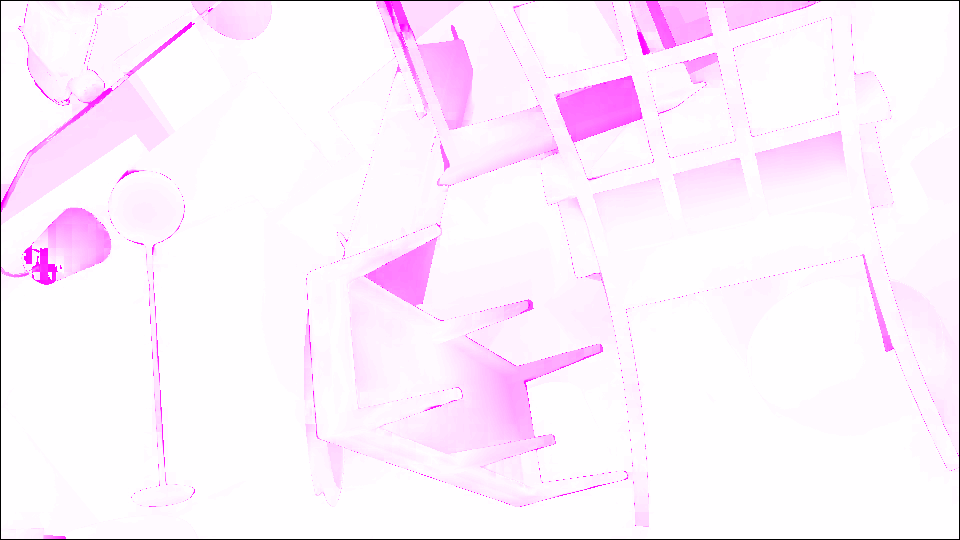}
    \put(10,-12){\color{black}\scriptsize$\rm AEPE^{all}_{2D}$=5.95}
    \end{overpic}
    \begin{overpic}[width=.12\linewidth]{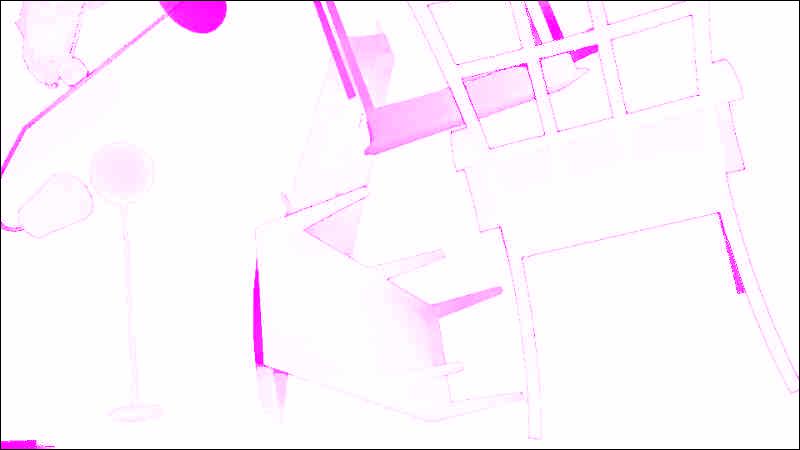}
    \put(10,-12){\color{black}\scriptsize$\rm AEPE^{all}_{2D}$=2.35}
    \end{overpic}
  \end{tabular}
\end{center}
\caption{
Examples of optical flow estimation error in the FlyingThings3D-clean dataset.
The more vivid the magenta, the higher the error.
{\mname}-2D method handles optical flow estimation better than RGB-based methods, while {\mname}-3D method outperforms RAFT-3D with a smaller AEPE. Best viewed in color.}
\label{fig:FlyingThings3D}
\end{figure*}
\begin{figure*}[t]
\vspace*{3mm}
\begin{center}
  \begin{tabular}{@{}c@{}c@{}c@{}c@{}c@{}c}
    \vspace*{3mm}
    \begin{overpic}[width=.12\linewidth]{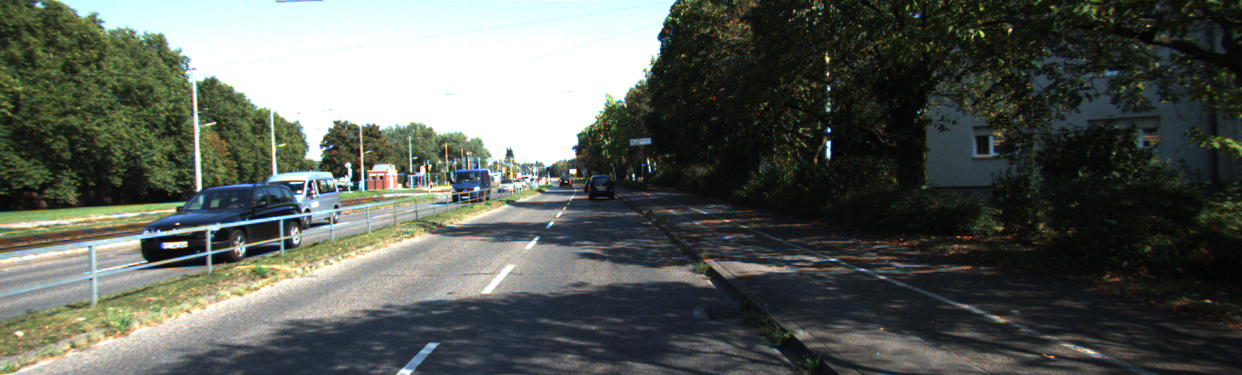}
     \put(38,-12){\color{black}\scriptsize\textbf{RGB}}
    \end{overpic}
    
    \begin{overpic}[width=.12\linewidth]{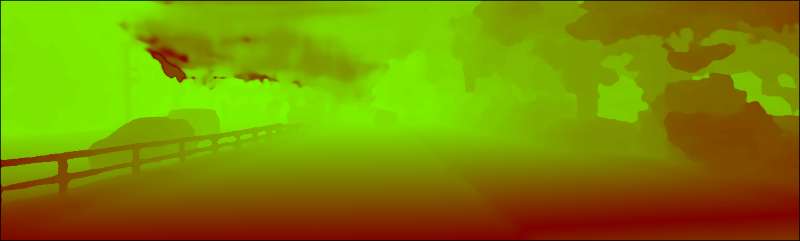}
    \put(38,-12){\color{black}\scriptsize\textbf{depth}}
    \end{overpic}
    \begin{overpic}[width=.12\linewidth]{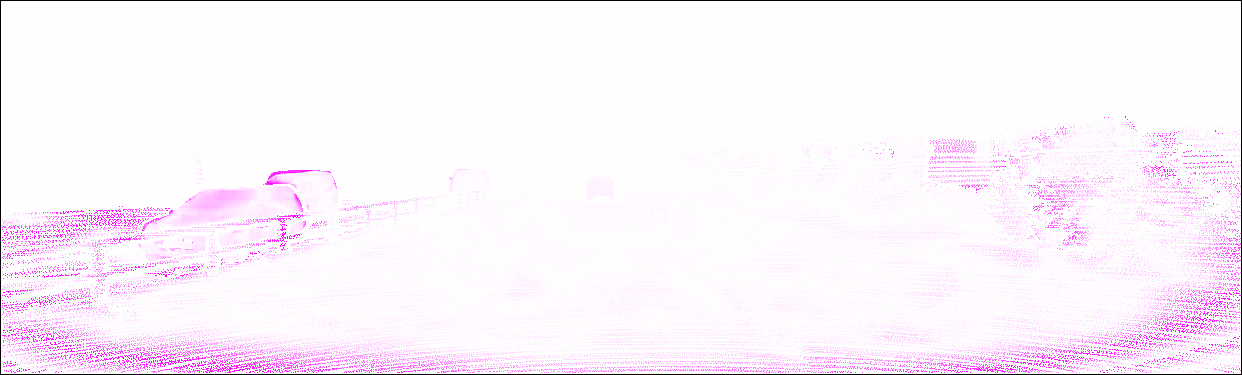}
    \put(38,35){\color{black}\scriptsize\textbf{RAFT}}
    \put(10,-12){\color{black}\scriptsize$\rm AEPE^{all}_{2D}$=4.99}
    \end{overpic}
    \begin{overpic}[width=.12\linewidth]{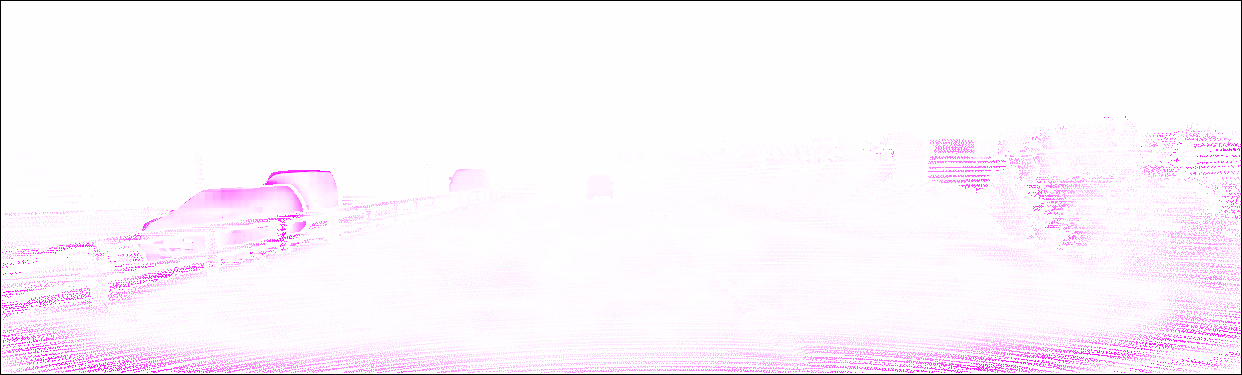}
    \put(36,35){\color{black}\scriptsize\textbf{GMA}}
    \put(10,-12){\color{black}\scriptsize$\rm AEPE^{all}_{2D}$=4.30}
     \end{overpic}
    \begin{overpic}[width=.12\linewidth]{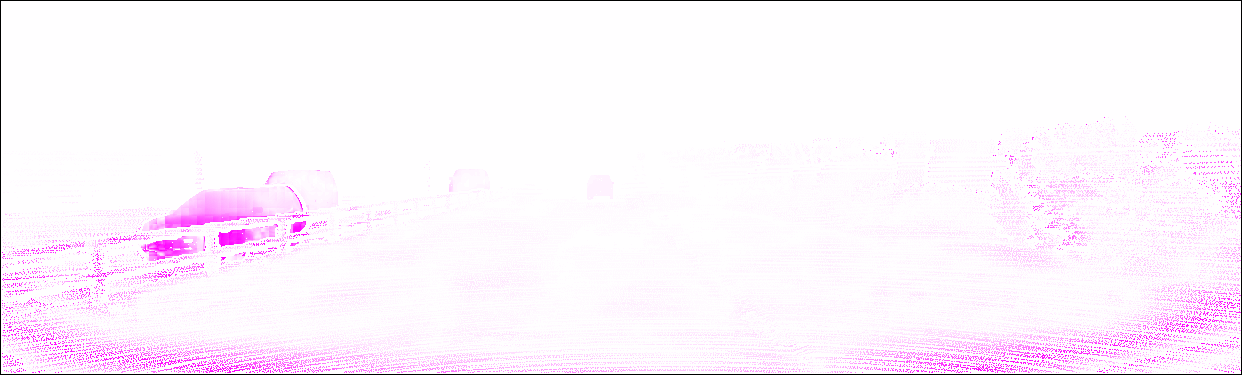}
    \put(10,35){\color{black}\scriptsize\textbf{{\mname}-2D}}
    \put(10,-12){\color{black}\scriptsize$\rm AEPE^{all}_{2D}$=4.11}
    \end{overpic}
    \begin{overpic}[width=.12\linewidth]{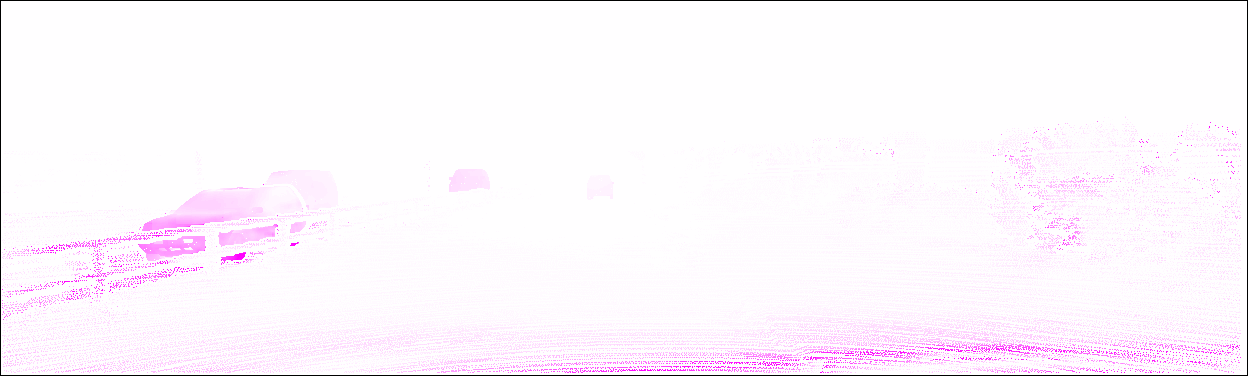}
    \put(29,35){\color{black}\scriptsize\textbf{RAFT-3D}}
    \put(10,-12){\color{black}\scriptsize$\rm AEPE^{all}_{2D}$=3.80}
    \end{overpic}
    \begin{overpic}[width=.12\linewidth]{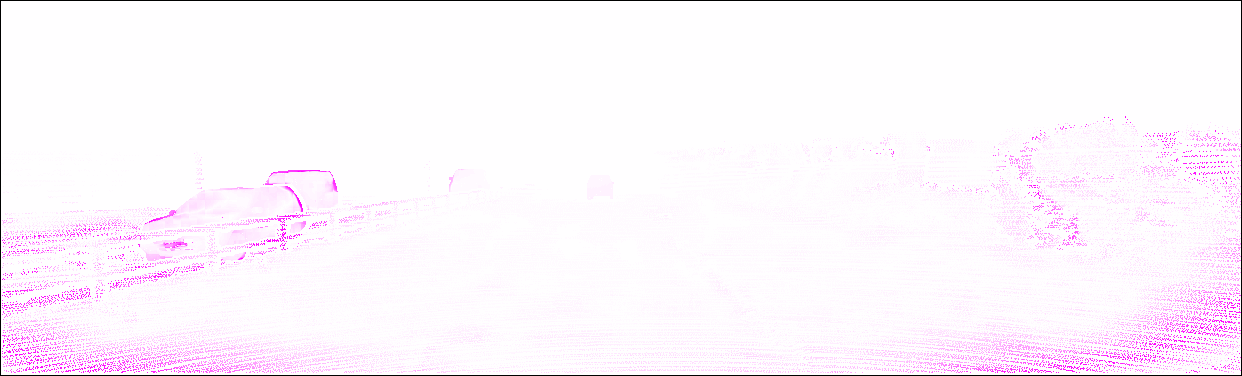}
    \put(19,35){\color{black}\scriptsize\textbf{CamLiRAFT}}
    \put(10,-12){\color{black}\scriptsize$\rm AEPE^{all}_{2D}$=3.49}
    \end{overpic}
    \begin{overpic}[width=.12\linewidth]{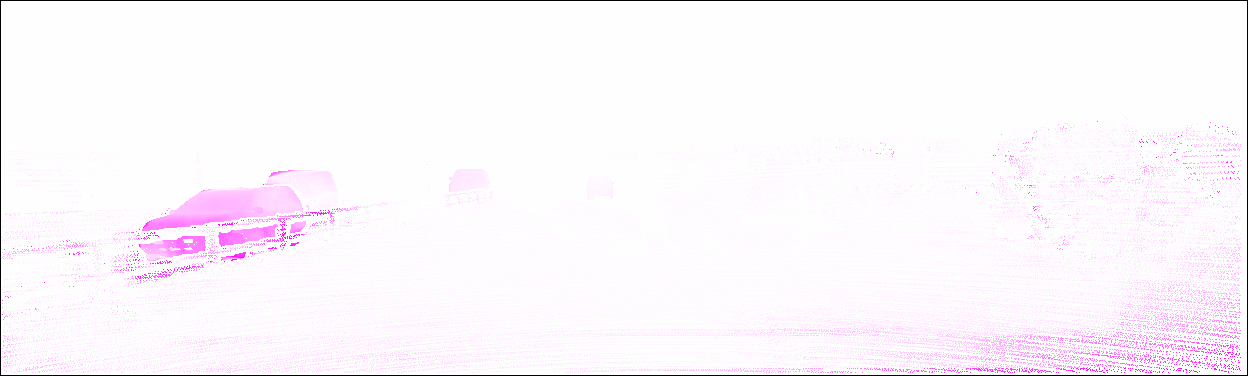}
    \put(10,35){\color{black}\scriptsize\textbf{{\mname}-3D}}
    \put(10,-12){\color{black}\scriptsize$\rm AEPE^{all}_{2D}$=2.51}
    \end{overpic}\\
    \vspace*{3mm}
    \begin{overpic}[width=.12\linewidth]{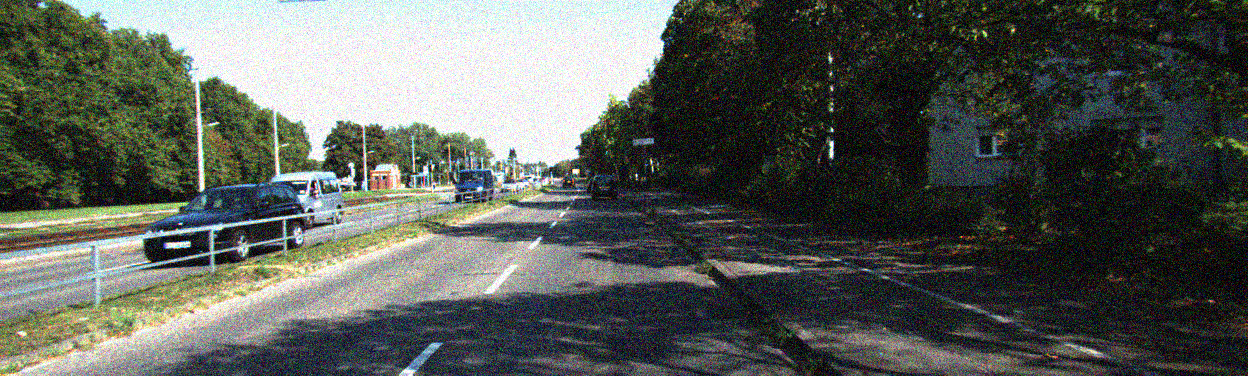}
     \put(38,-12){\color{black}\scriptsize\textbf{RGB}}
    \end{overpic}
    
    \begin{overpic}[width=.12\linewidth]{sections/images/flow-error-kitti/depthkitti.jpg}
    \put(38,-12){\color{black}\scriptsize\textbf{depth}}
    \end{overpic}
    \begin{overpic}[width=.12\linewidth]{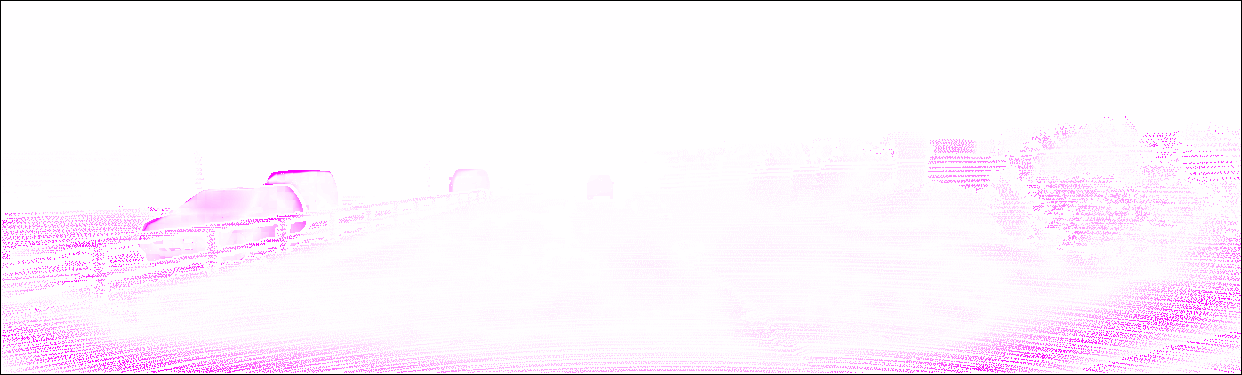}
    \put(10,-12){\color{black}\scriptsize$\rm AEPE^{all}_{2D}$=5.93}
    \end{overpic}
    \begin{overpic}[width=.12\linewidth]{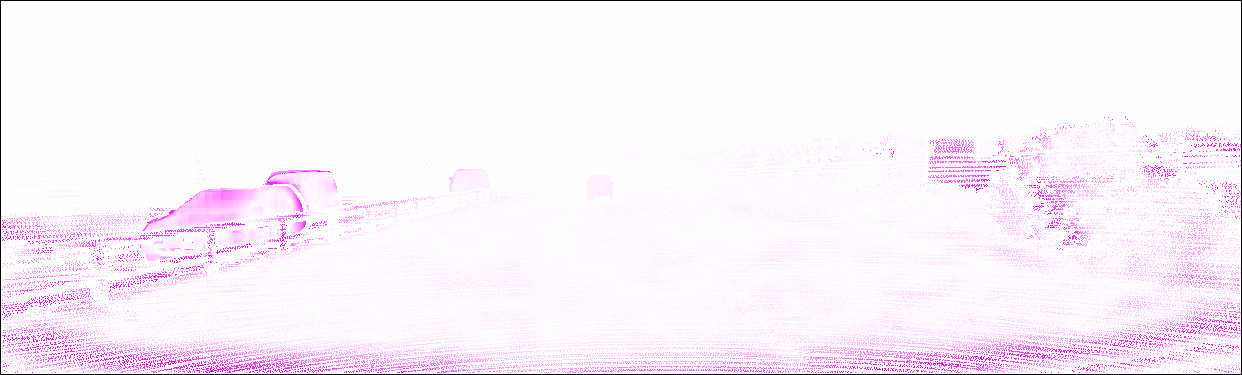}
    \put(10,-12){\color{black}\scriptsize$\rm AEPE^{all}_{2D}$=4.75}
     \end{overpic}
    \begin{overpic}[width=.12\linewidth]{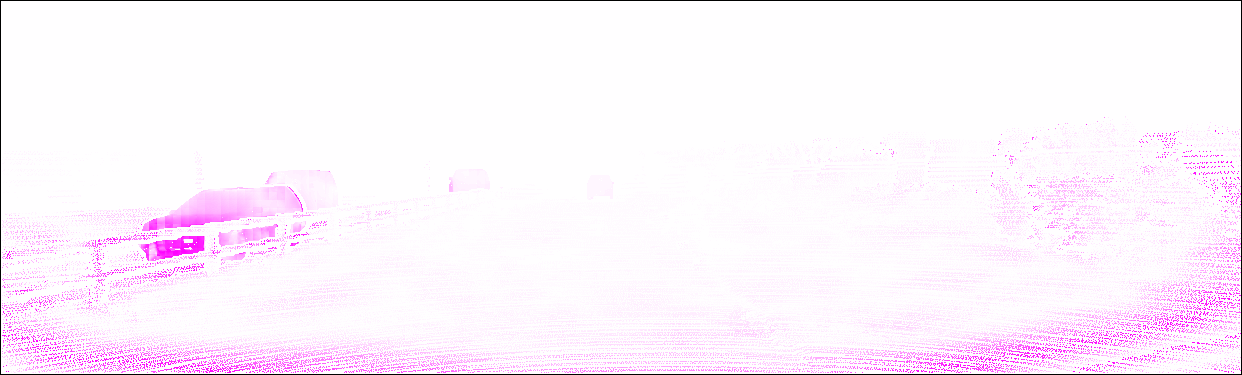}
    \put(10,-12){\color{black}\scriptsize$\rm AEPE^{all}_{2D}$=4.74}
    \end{overpic}
    \begin{overpic}[width=.12\linewidth]{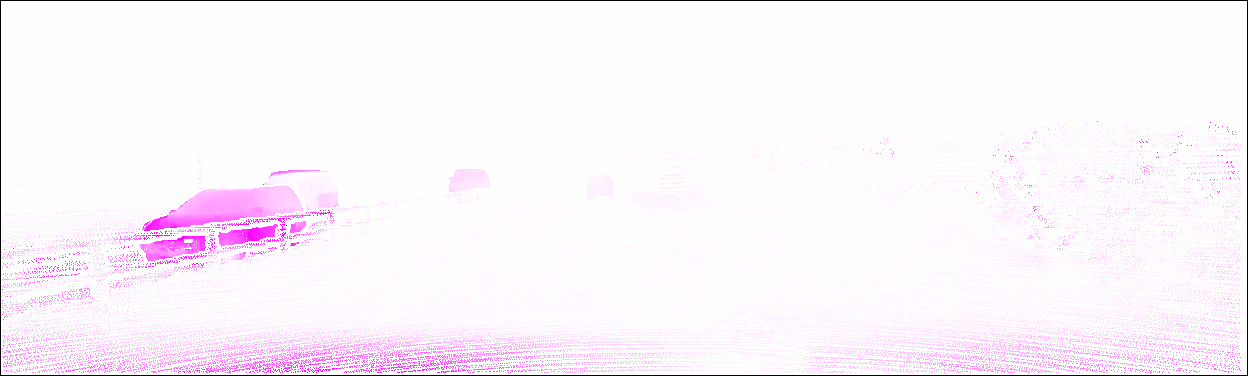}
    \put(10,-12){\color{black}\scriptsize$\rm AEPE^{all}_{2D}$=3.09}
    \end{overpic}
    \begin{overpic}[width=.12\linewidth]{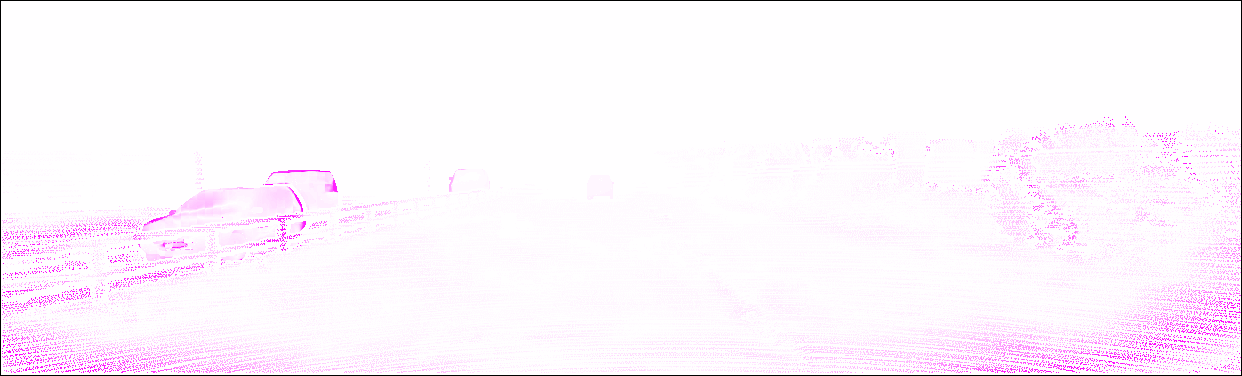}
    \put(10,-12){\color{black}\scriptsize$\rm AEPE^{all}_{2D}$=3.60}
    \end{overpic}
    \begin{overpic}[width=.12\linewidth]{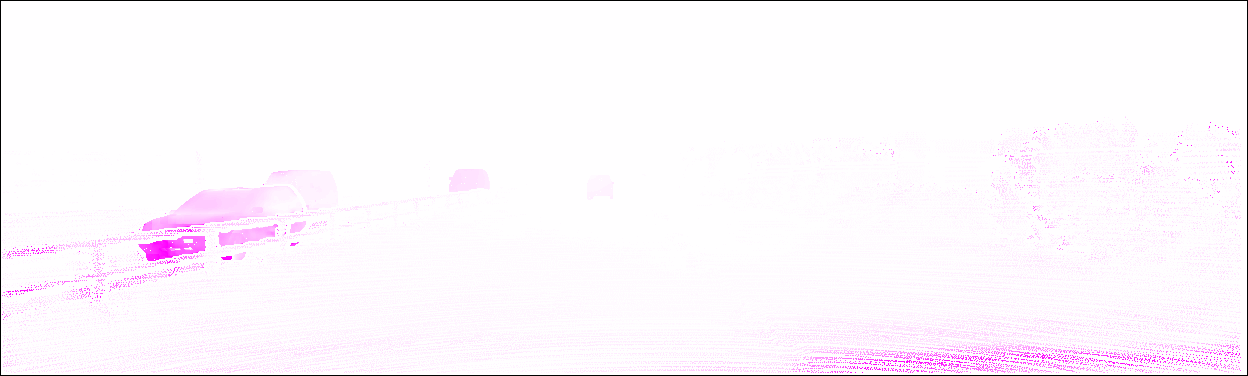}
    \put(10,-12){\color{black}\scriptsize$\rm AEPE^{all}_{2D}$=2.70}
    \end{overpic}\\
    \begin{overpic}[width=.12\linewidth]{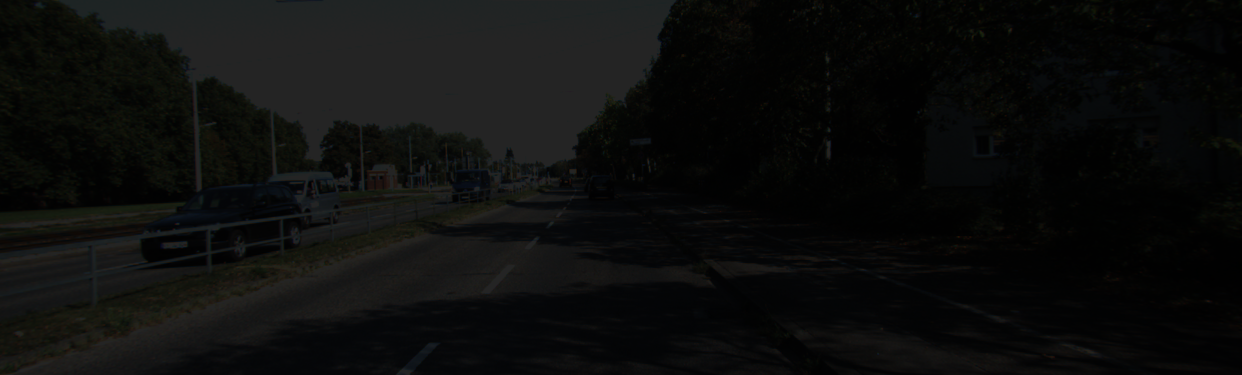}
     \put(38,-12){\color{black}\scriptsize\textbf{RGB}}
    \end{overpic}
    \begin{overpic}[width=.12\linewidth]{sections/images/flow-error-kitti/depthkitti.jpg}
    \put(38,-12){\color{black}\scriptsize\textbf{depth}}
    \end{overpic}
    \begin{overpic}[width=.12\linewidth]{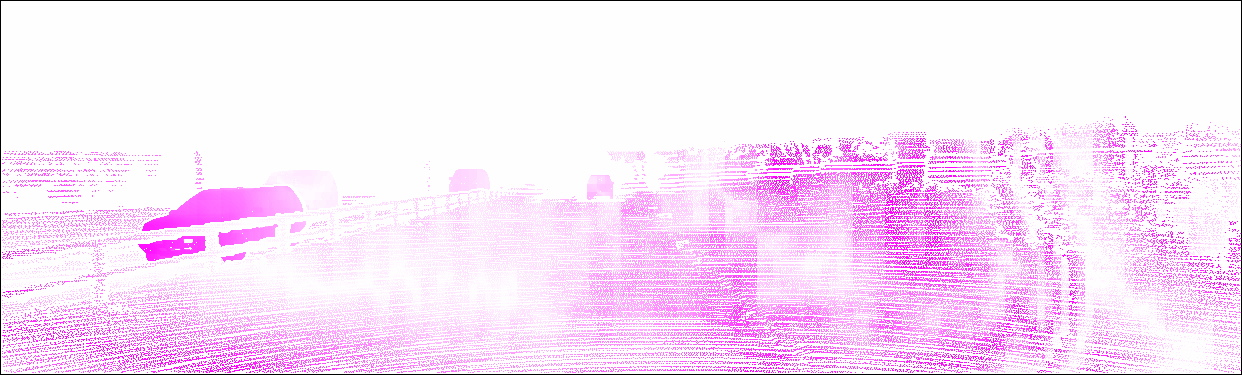}
    \put(10,-12){\color{black}\scriptsize$\rm AEPE^{all}_{2D}$=52.98}
    \end{overpic}
    \begin{overpic}[width=.12\linewidth]{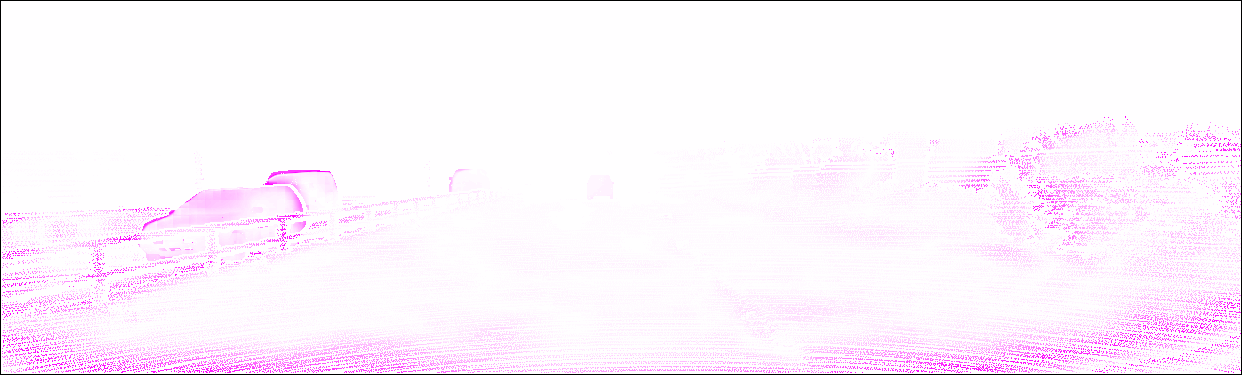}
    \put(10,-12){\color{black}\scriptsize$\rm AEPE^{all}_{2D}$=5.52}
     \end{overpic}
    \begin{overpic}[width=.12\linewidth]{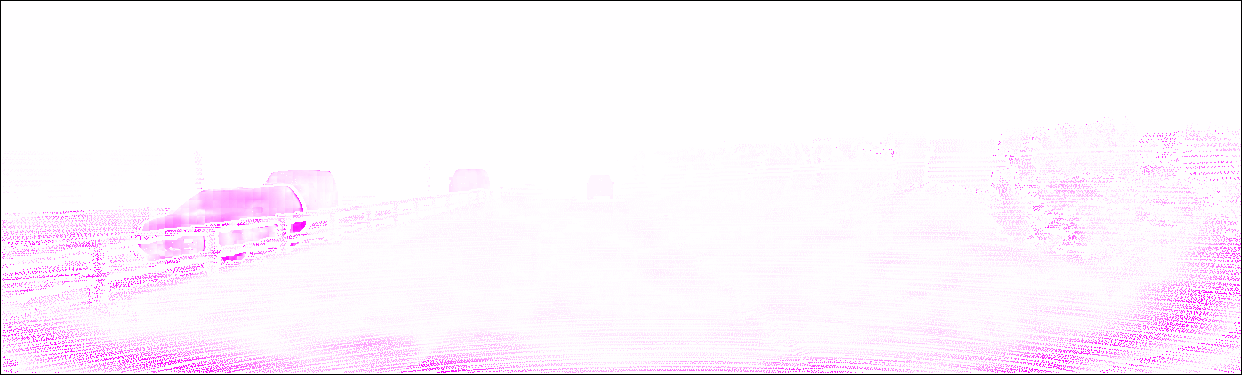}
    \put(10,-12){\color{black}\scriptsize$\rm AEPE^{all}_{2D}$=4.82}
    \end{overpic}
    \begin{overpic}[width=.12\linewidth]{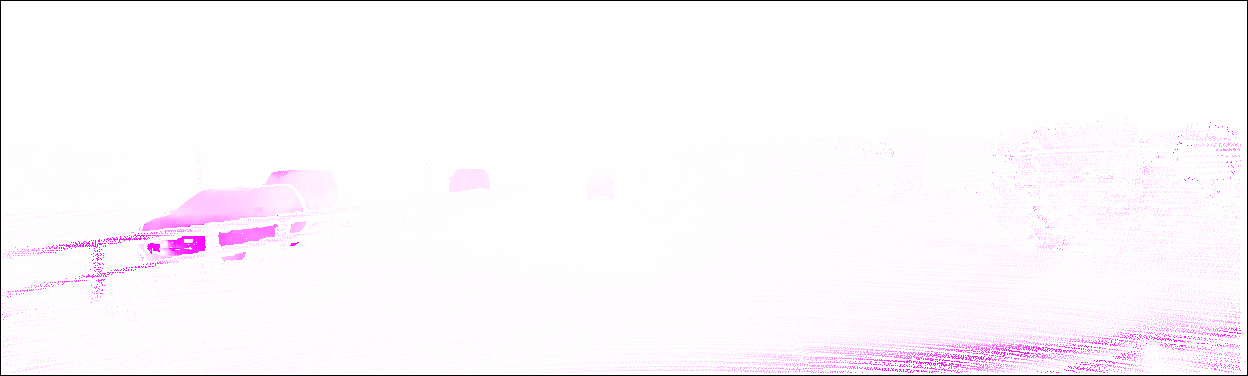}
    \put(10,-12){\color{black}\scriptsize$\rm AEPE^{all}_{2D}$=8.75}
    \end{overpic}
    \begin{overpic}[width=.12\linewidth]{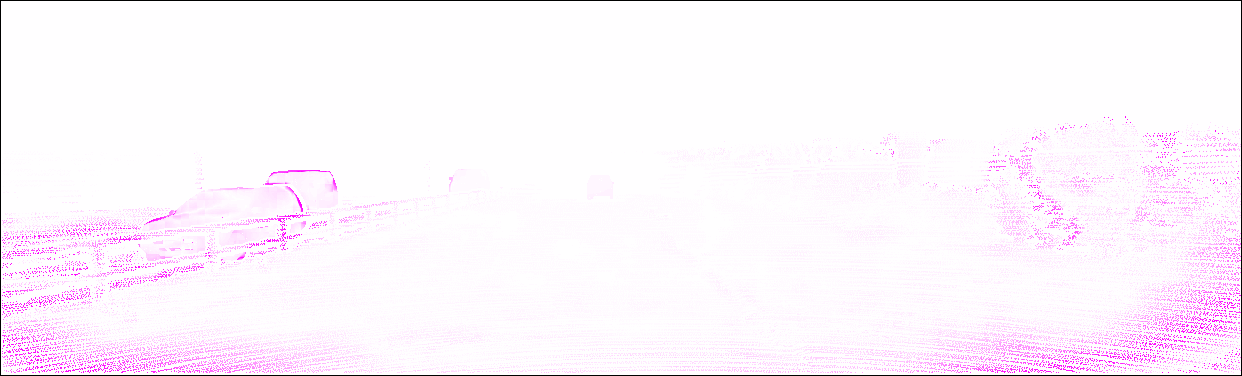}
    \put(10,-12){\color{black}\scriptsize$\rm AEPE^{all}_{2D}$=3.74}
    \end{overpic}
    \begin{overpic}[width=.12\linewidth]{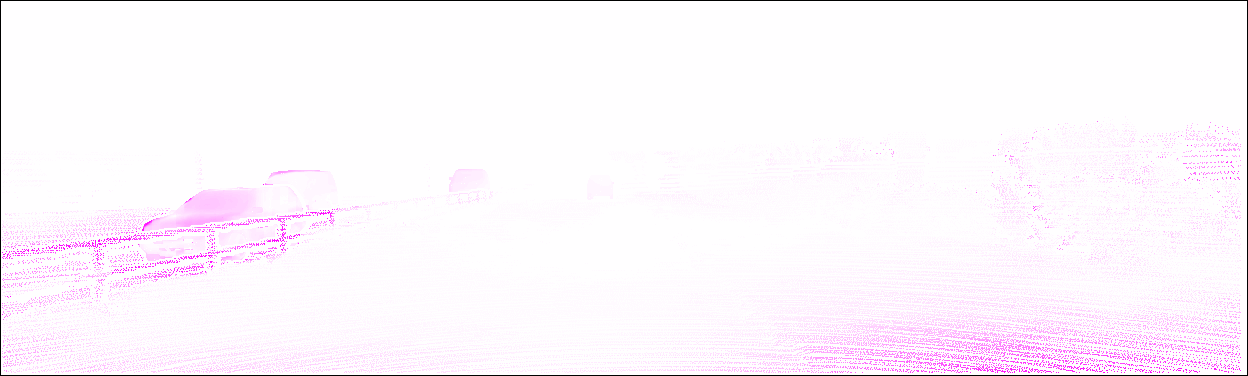}
    \put(10,-12){\color{black}\scriptsize$\rm AEPE^{all}_{2D}$=3.05}
    \end{overpic}\\
  \end{tabular}
\end{center}
\caption{Examples of optical flow estimation error in the KITTI dataset.
The more vivid the magenta, the higher the error.
{\mname}-2D method handles optical flow estimation better than all RGB-based methods. 
{\mname}-3D method outperforms RAFT-3D with a smaller AEPE. Best viewed in color.}
\label{fig:KITTI}
\end{figure*}
\begin{figure*}[h]
\begin{center}
  \begin{tabular}{@{}c@{}c@{}c@{}c@{}c@{}c}
    \vspace*{3mm}
    \begin{overpic}[width=.12\linewidth]{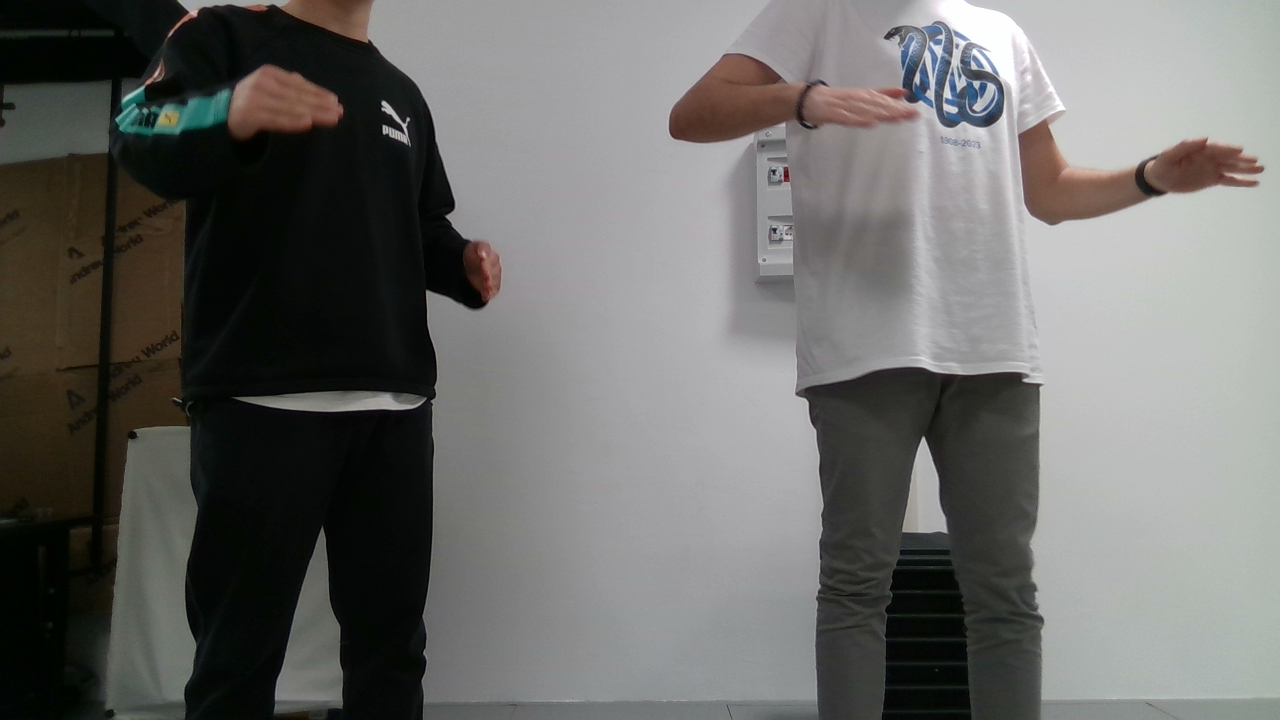}
     \put(38,-12){\color{black}\scriptsize\textbf{RGB}}
    \end{overpic}
    
    \begin{overpic}[width=.12\linewidth]{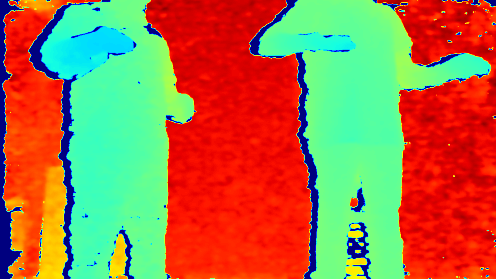}
    \put(38,-12){\color{black}\scriptsize\textbf{depth}}
    \end{overpic}
    \begin{overpic}[width=.12\linewidth]{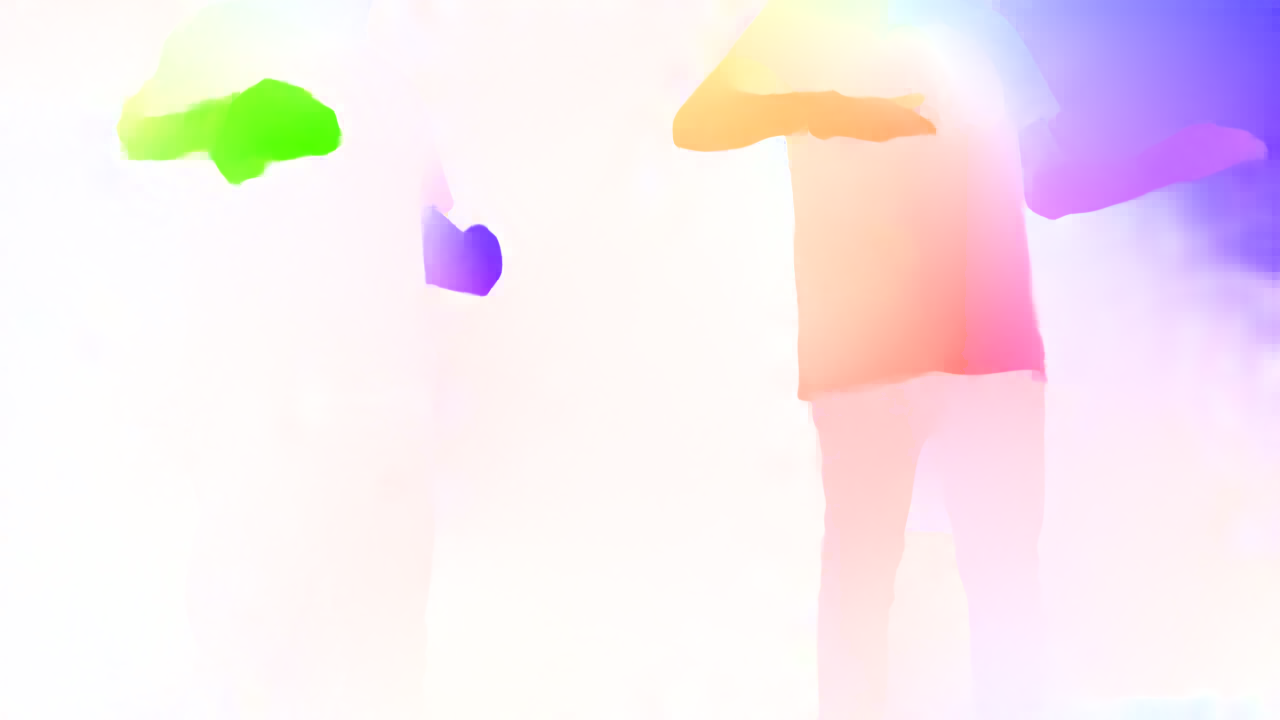}
    \put(38,60){\color{black}\scriptsize\textbf{RAFT}}
    \end{overpic}
    \begin{overpic}[width=.12\linewidth]{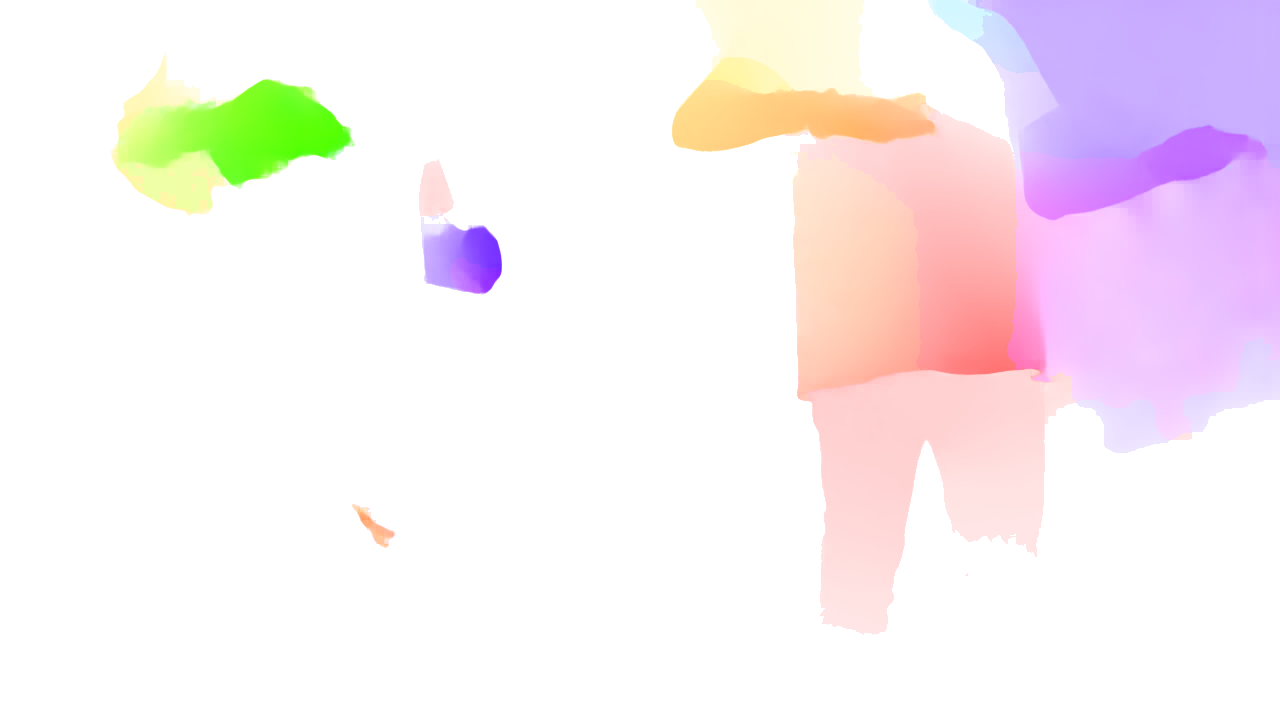}
    \put(36,60){\color{black}\scriptsize\textbf{GMA}}
     \end{overpic}
    \begin{overpic}[width=.12\linewidth]{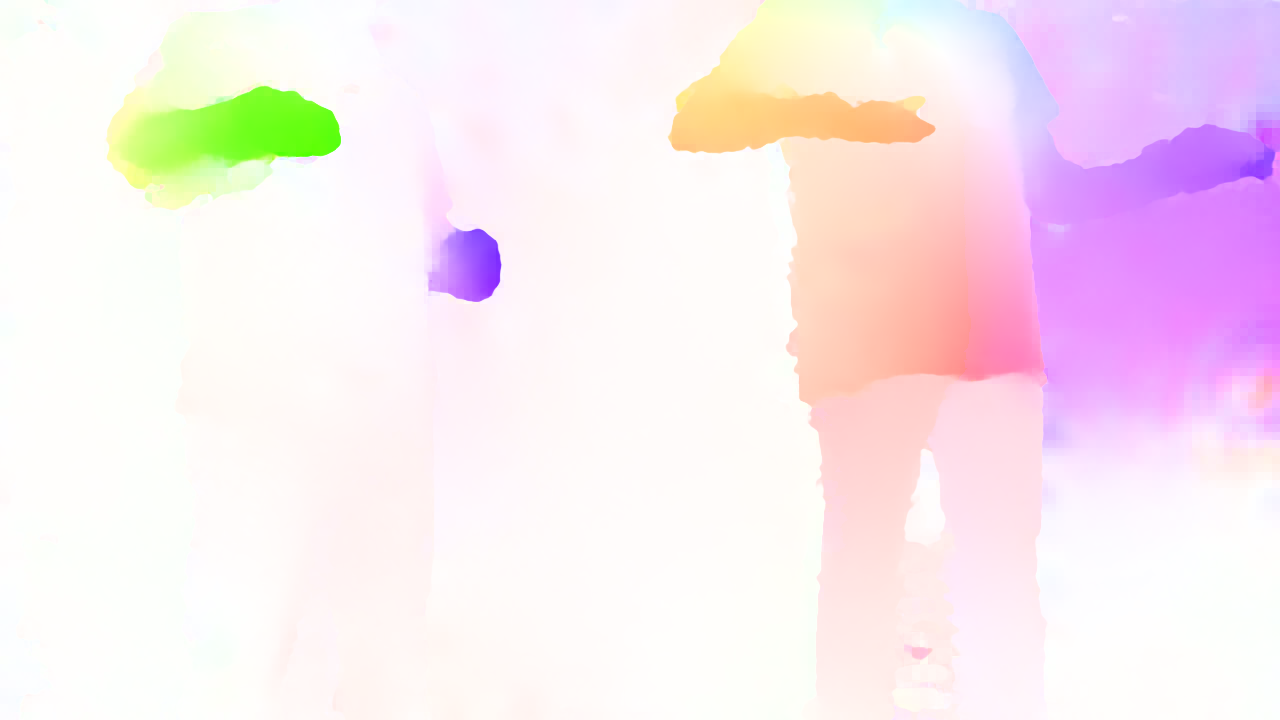}
    \put(10,60){\color{black}\scriptsize\textbf{{\mname}-2D}}
    \end{overpic}
    \begin{overpic}[width=.12\linewidth]{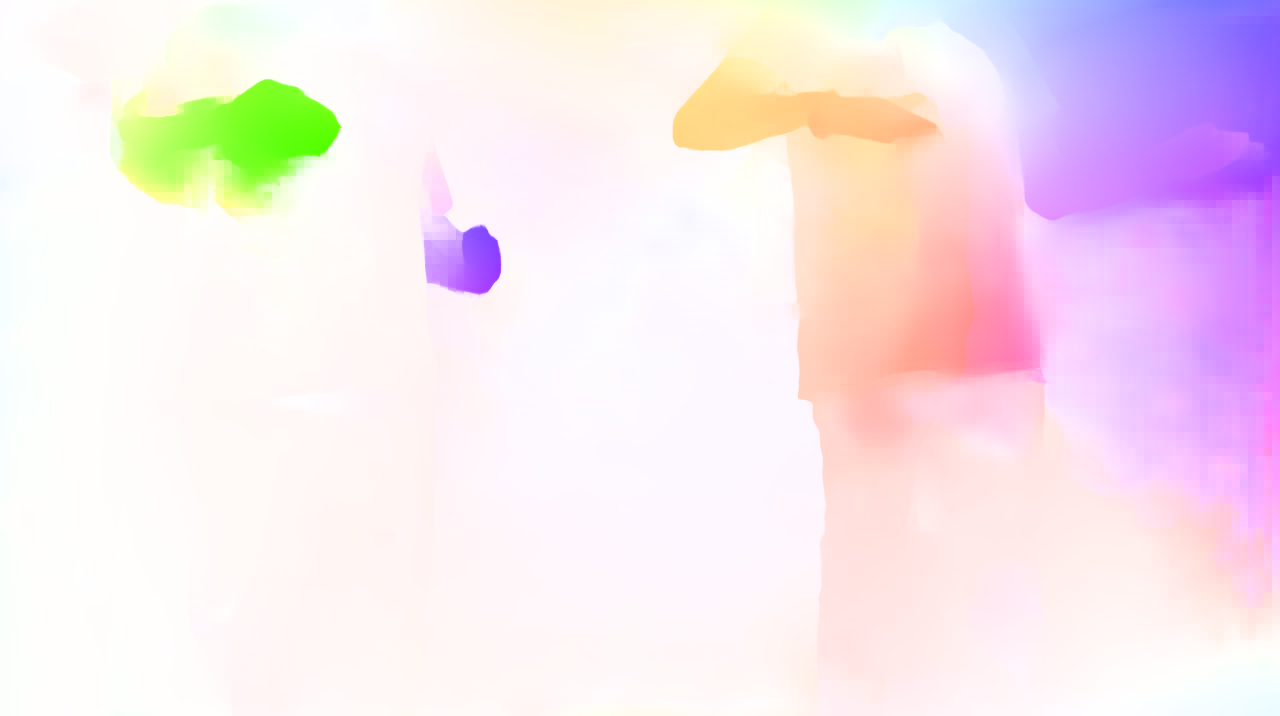}
    \put(29,60){\color{black}\scriptsize\textbf{RAFT-3D}}
    \end{overpic}
    \begin{overpic}[width=.12\linewidth]{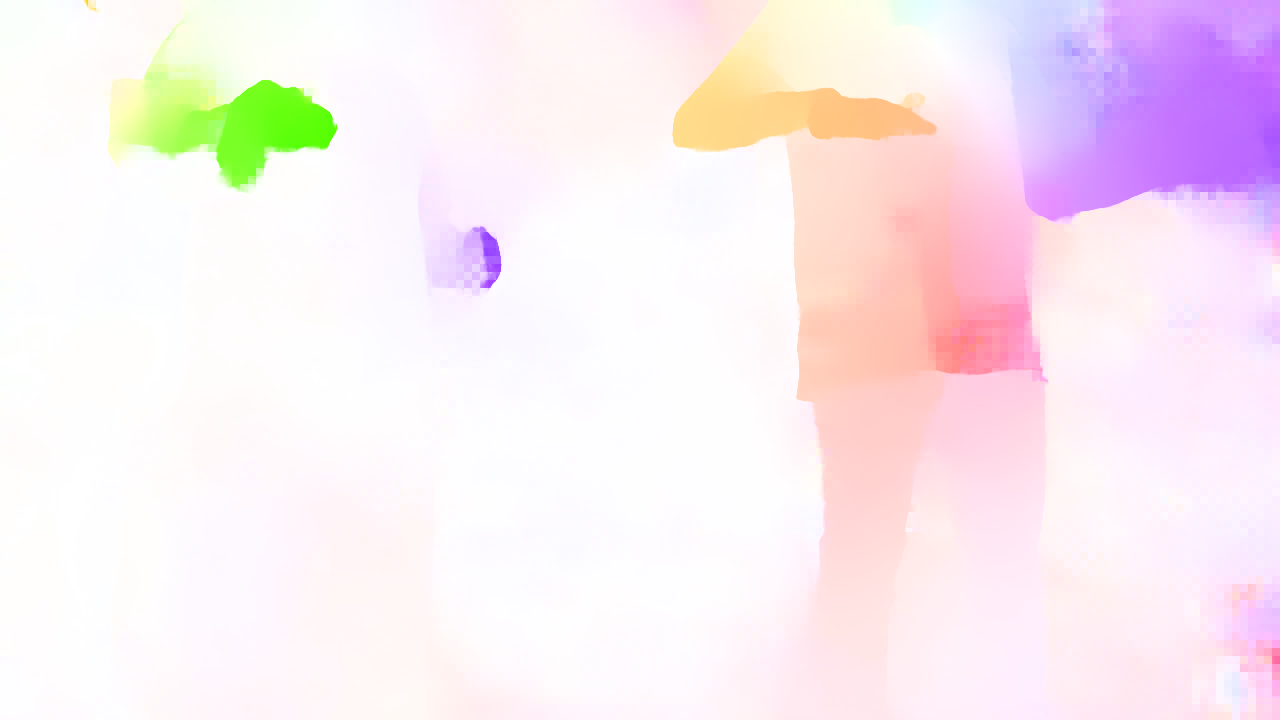}
    \put(22,60){\color{black}\scriptsize\textbf{CamLiRAFT}}
    \end{overpic}
    \begin{overpic}[width=.12\linewidth]{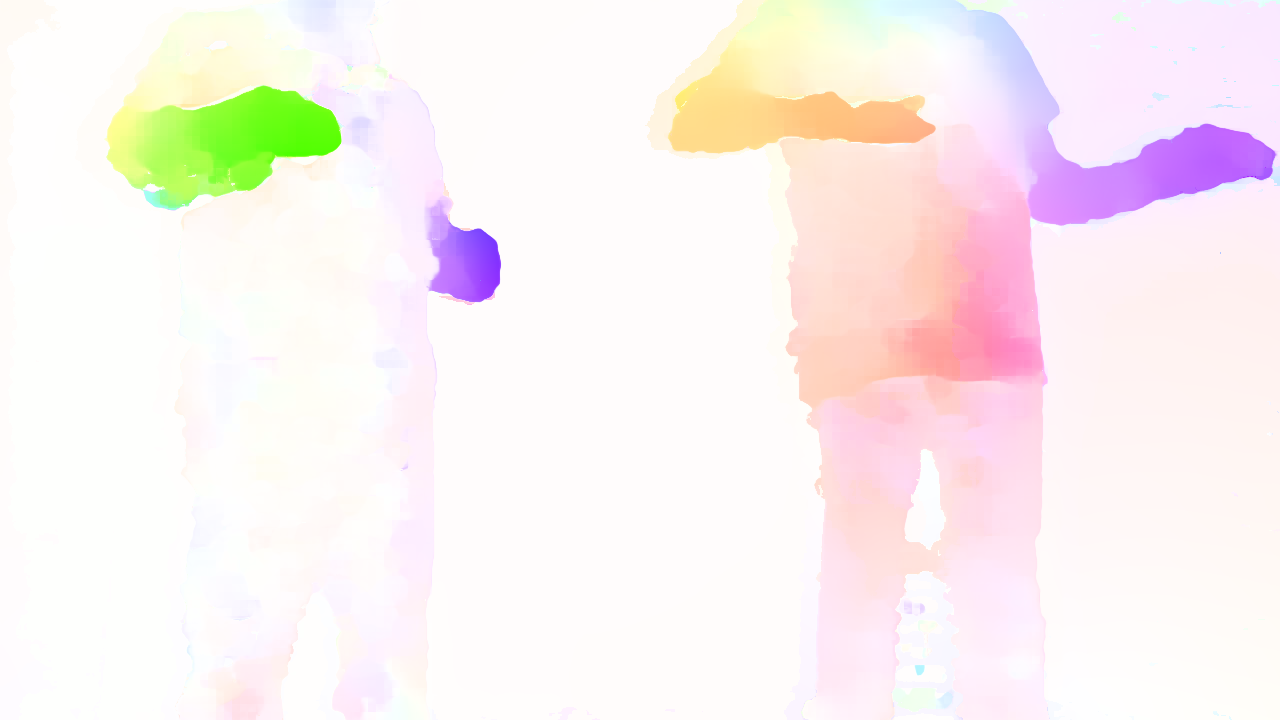}
    \put(10,60){\color{black}\scriptsize\textbf{{\mname}-3D}}
    \end{overpic}\\

    \vspace*{3mm}
    \begin{overpic}[width=.12\linewidth]{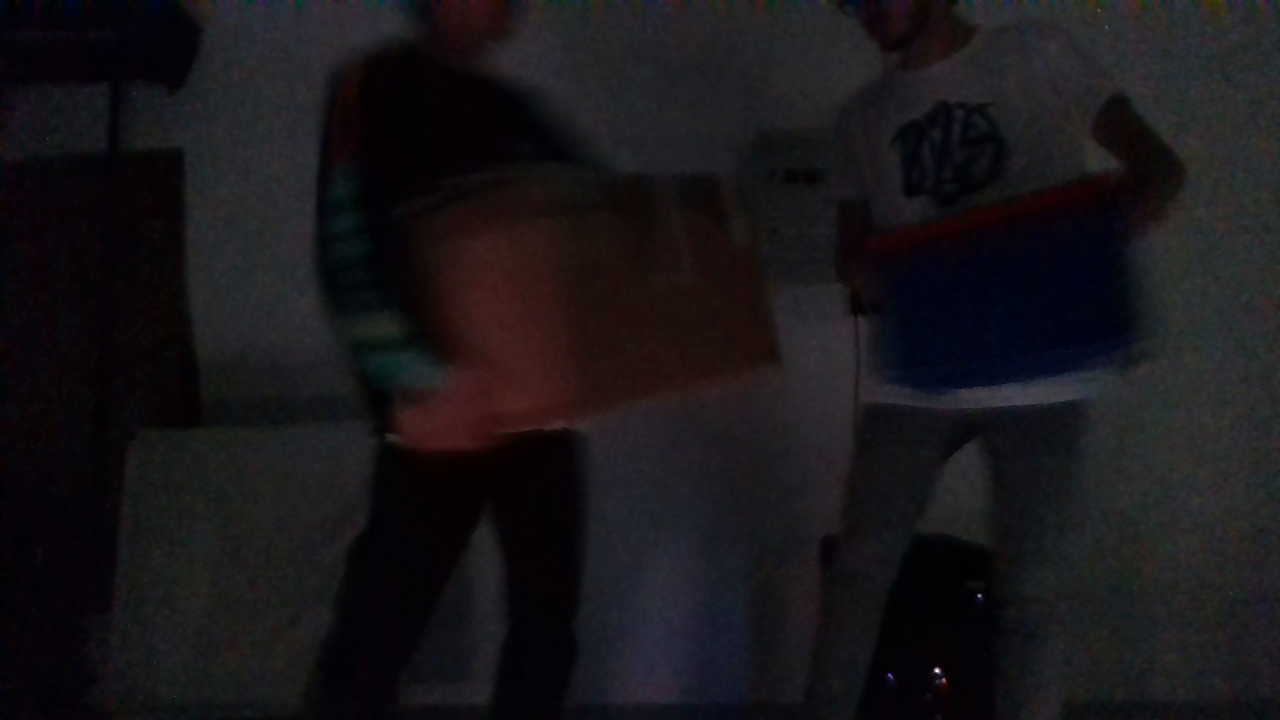}
     \put(38,-12){\color{black}\scriptsize\textbf{RGB}}
    \end{overpic}
    
    \begin{overpic}[width=.12\linewidth]{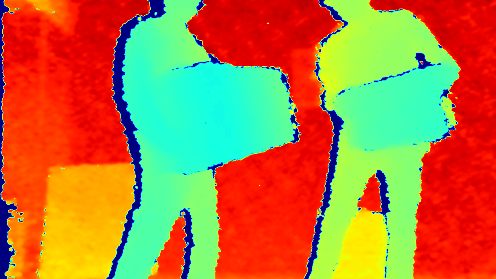}
    \put(38,-12){\color{black}\scriptsize\textbf{depth}}
    \end{overpic}
    \begin{overpic}[width=.12\linewidth]{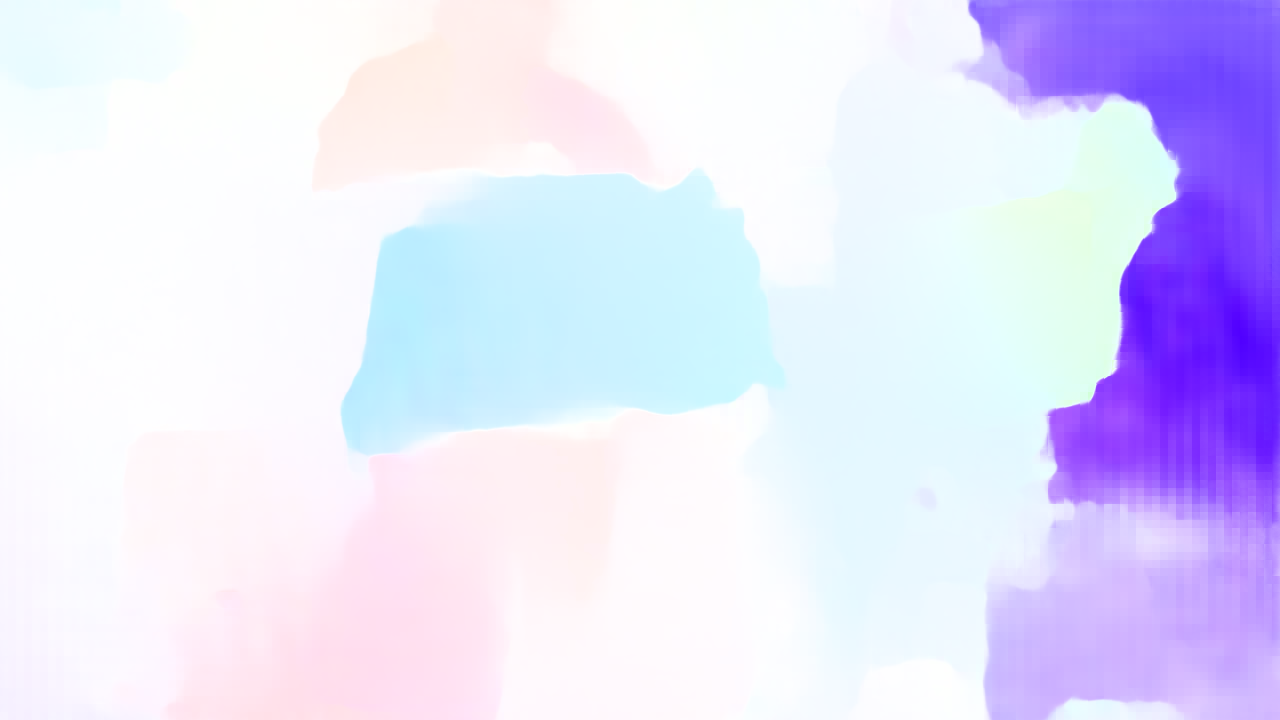}
    \end{overpic}
    \begin{overpic}[width=.12\linewidth]{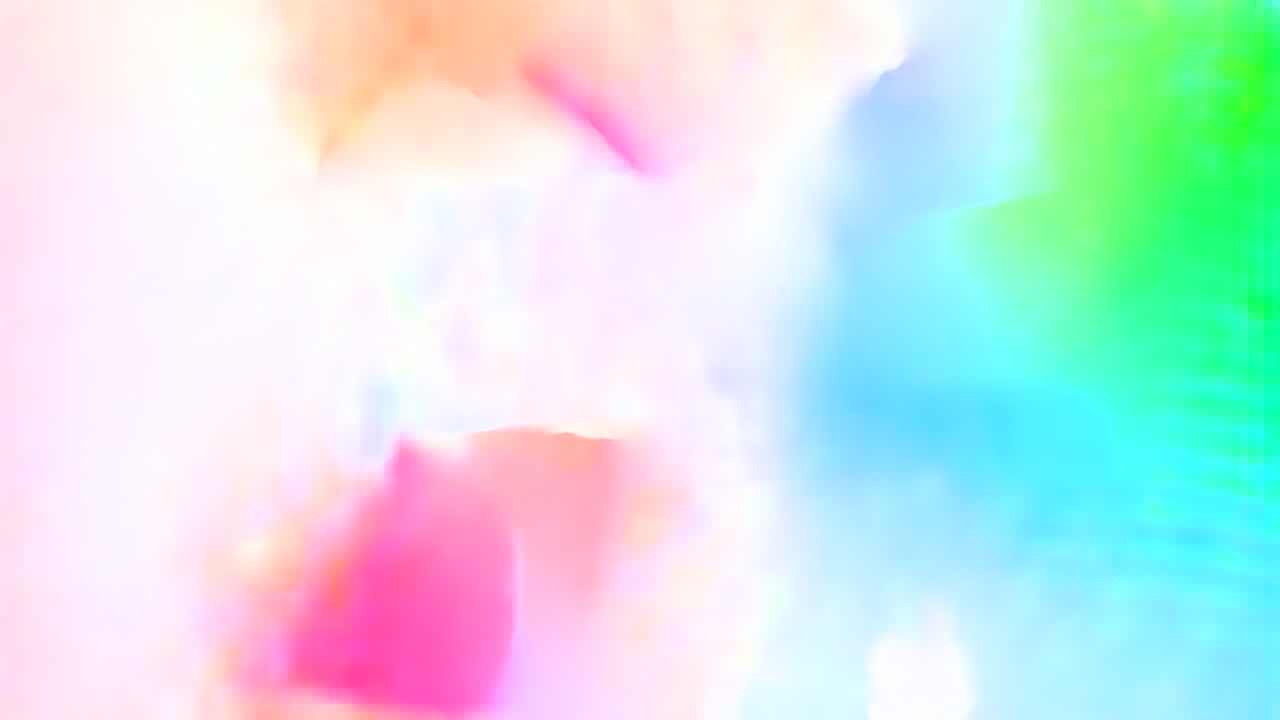}
     \end{overpic}
    \begin{overpic}[width=.12\linewidth]{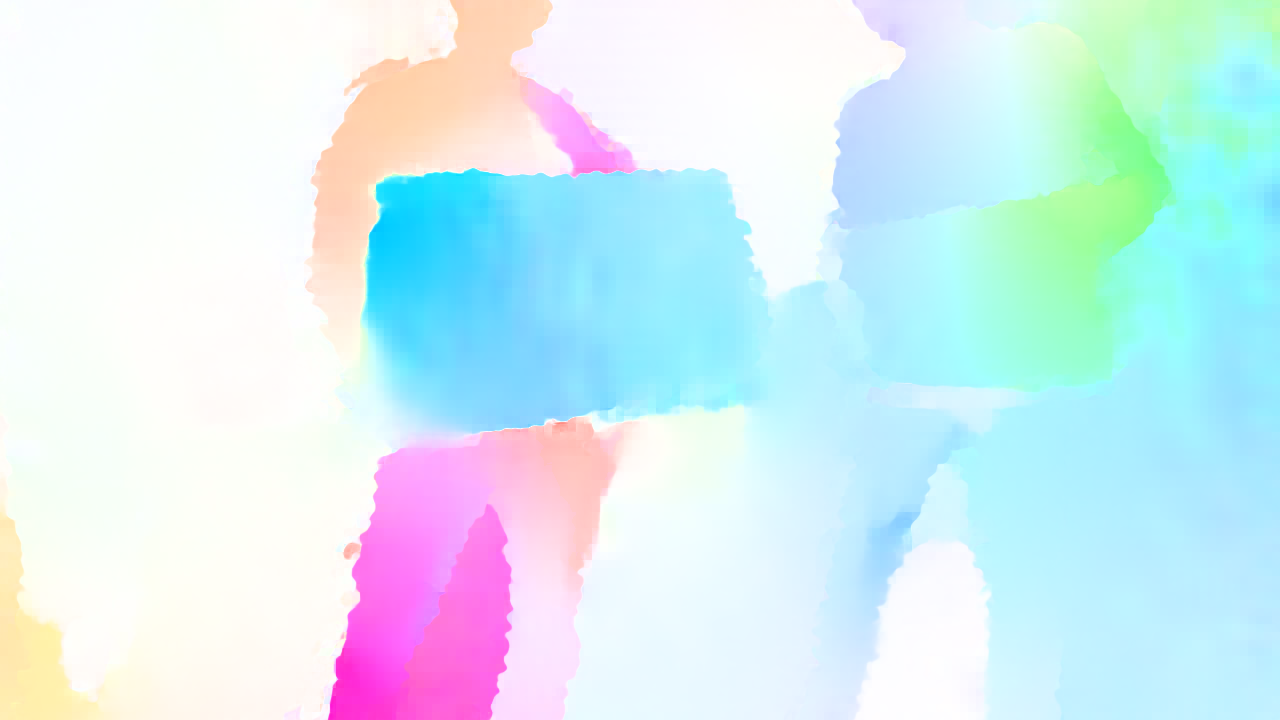}
    \end{overpic}
    \begin{overpic}[width=.12\linewidth]{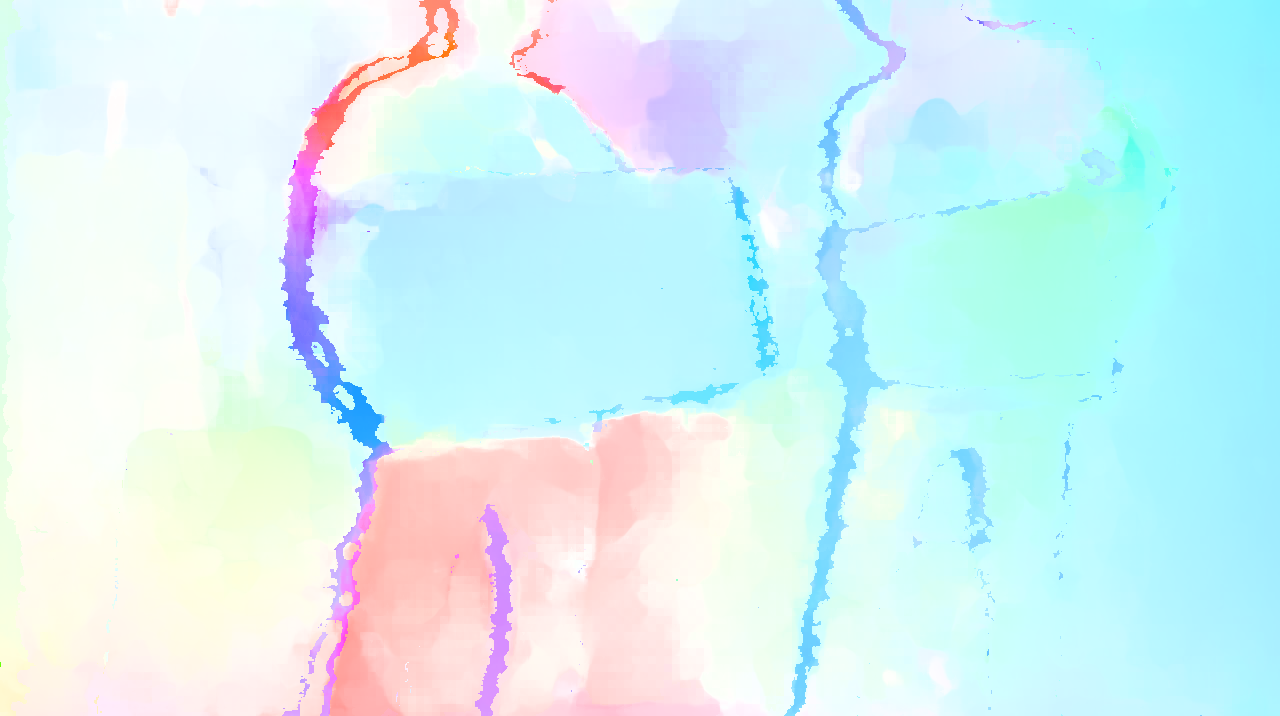}
    \end{overpic}
    \begin{overpic}[width=.12\linewidth]{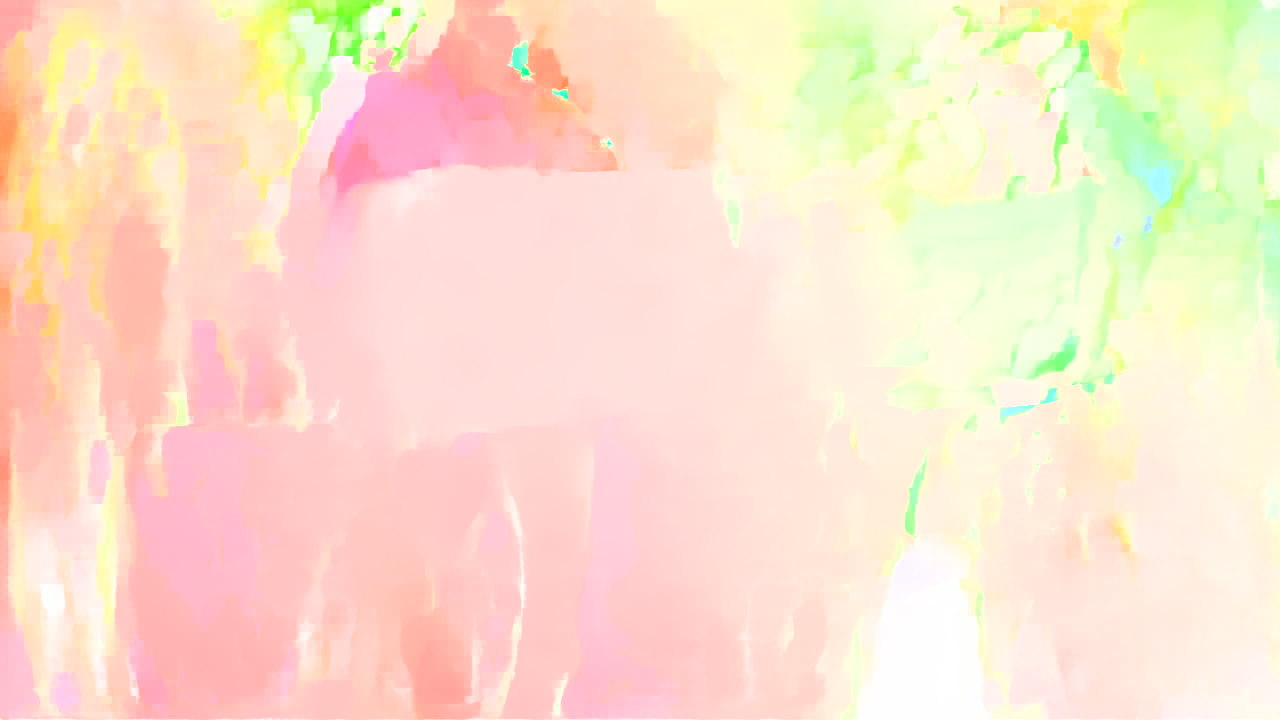}
    \end{overpic}
    \begin{overpic}[width=.12\linewidth]{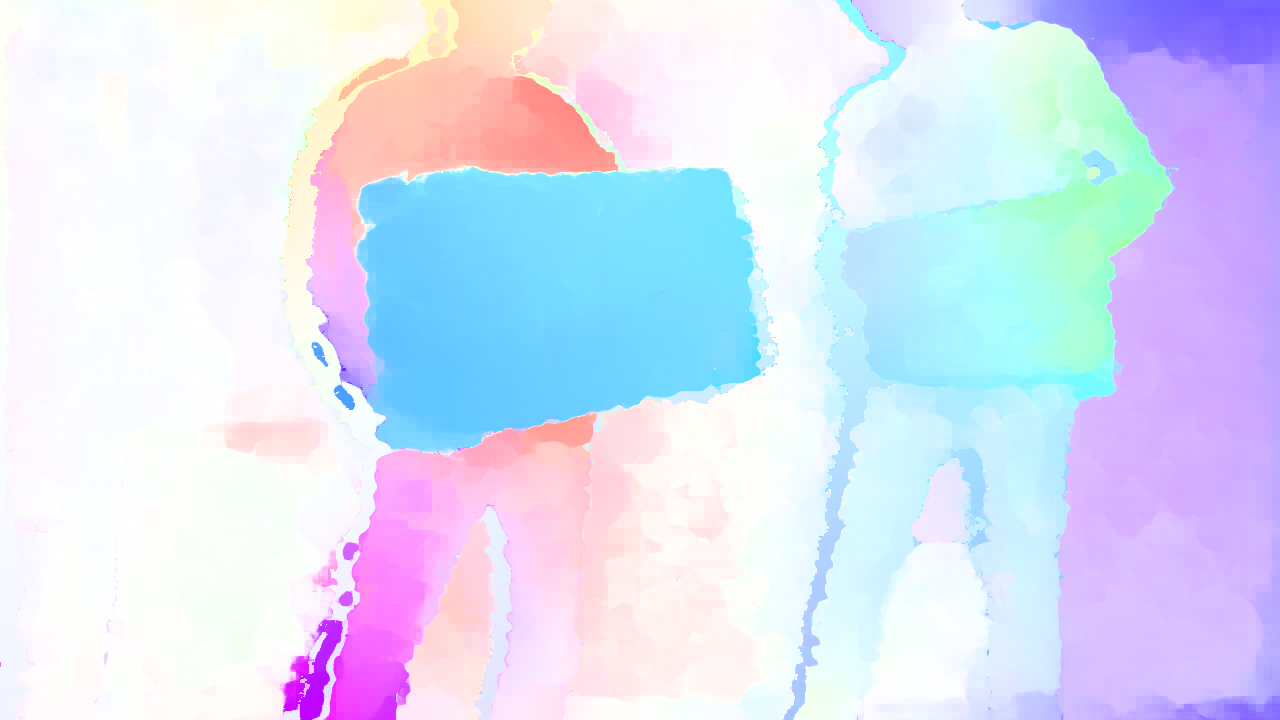}
    \end{overpic}\\
    \vspace*{0mm}
    \begin{overpic}[width=.12\linewidth]{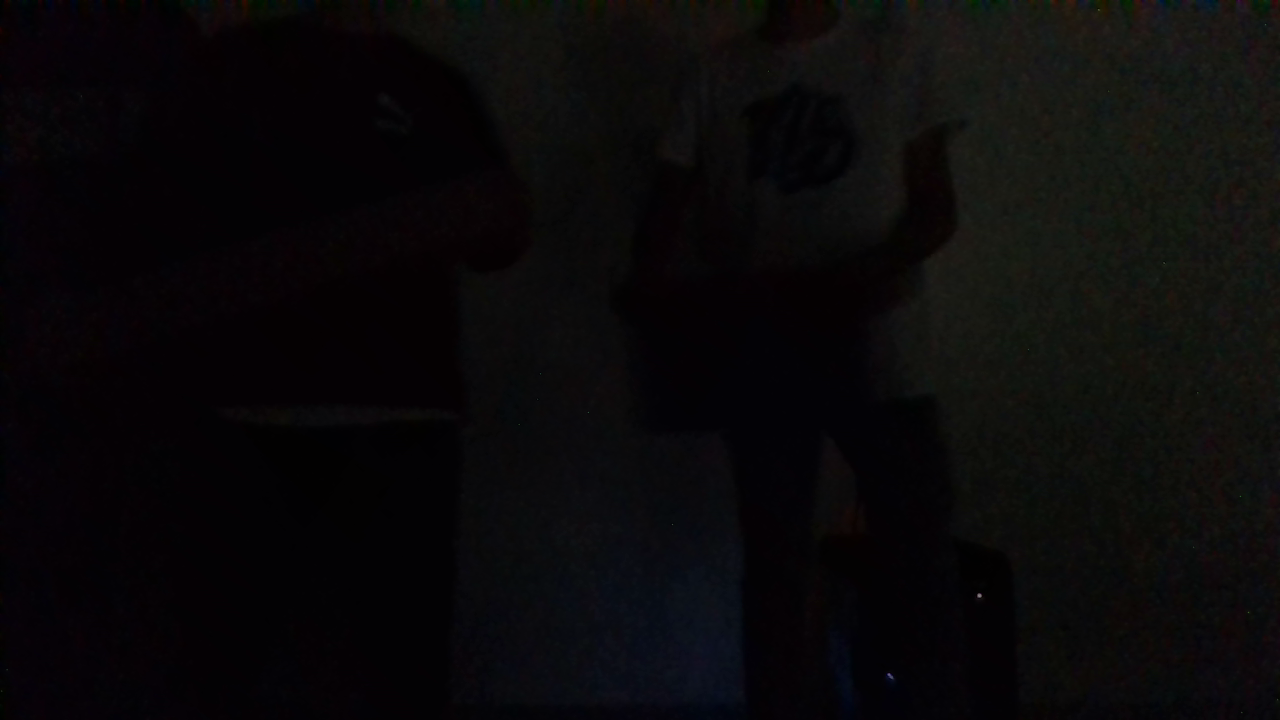}
     \put(38,-12){\color{black}\scriptsize\textbf{RGB}}
    \end{overpic}
    
    \begin{overpic}[width=.12\linewidth]{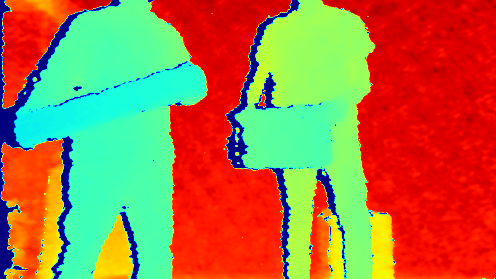}
    \put(38,-12){\color{black}\scriptsize\textbf{depth}}
    \end{overpic}
    \begin{overpic}[width=.12\linewidth]{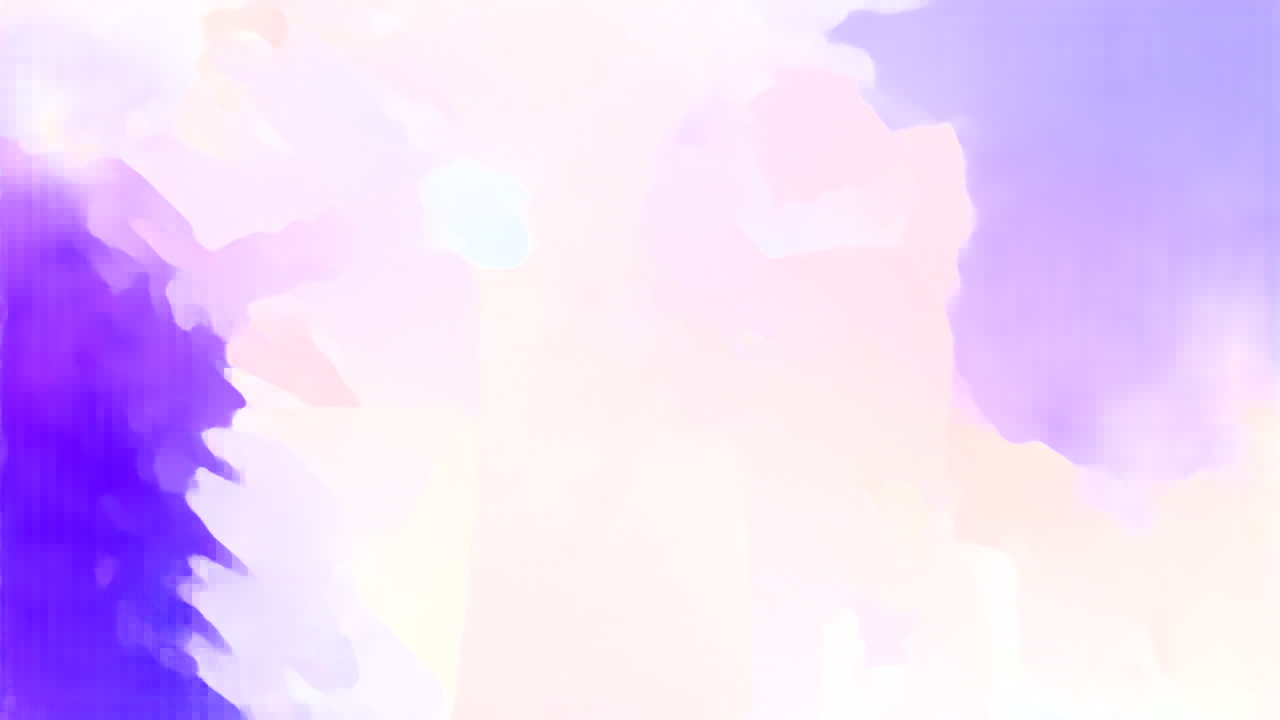}
    \end{overpic}
    \begin{overpic}[width=.12\linewidth]{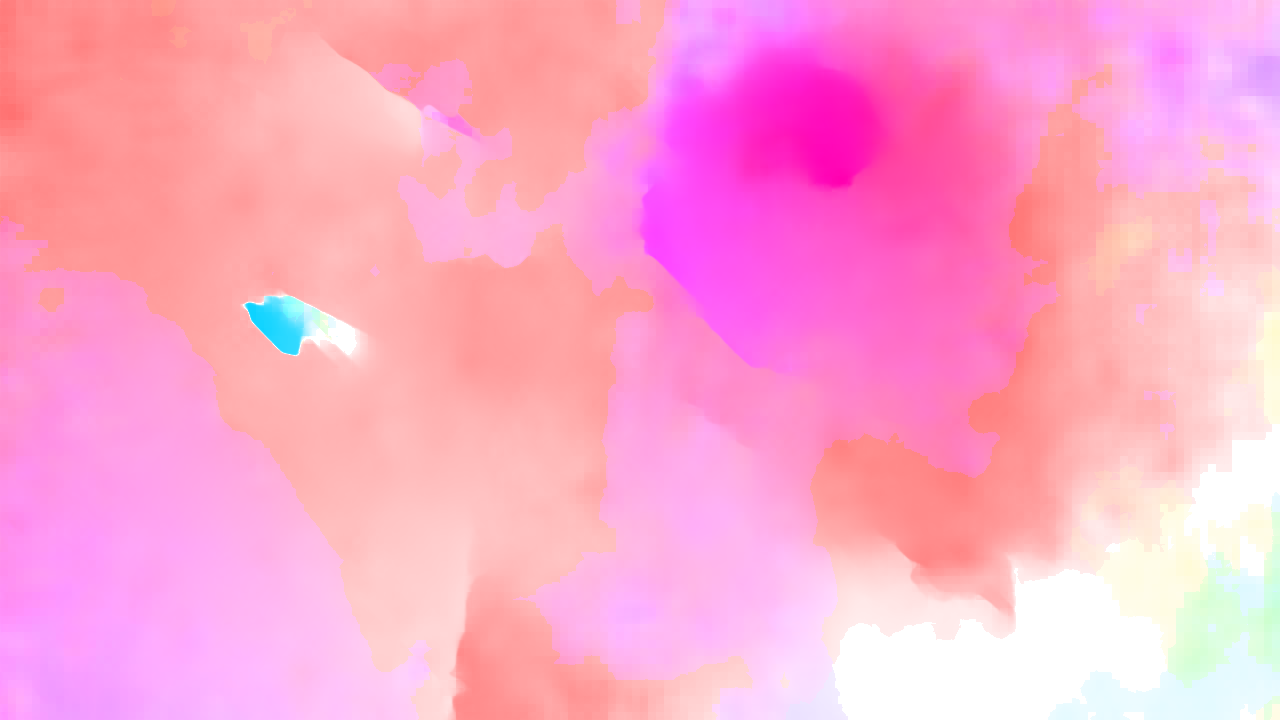}
     \end{overpic}
    \begin{overpic}[width=.12\linewidth]{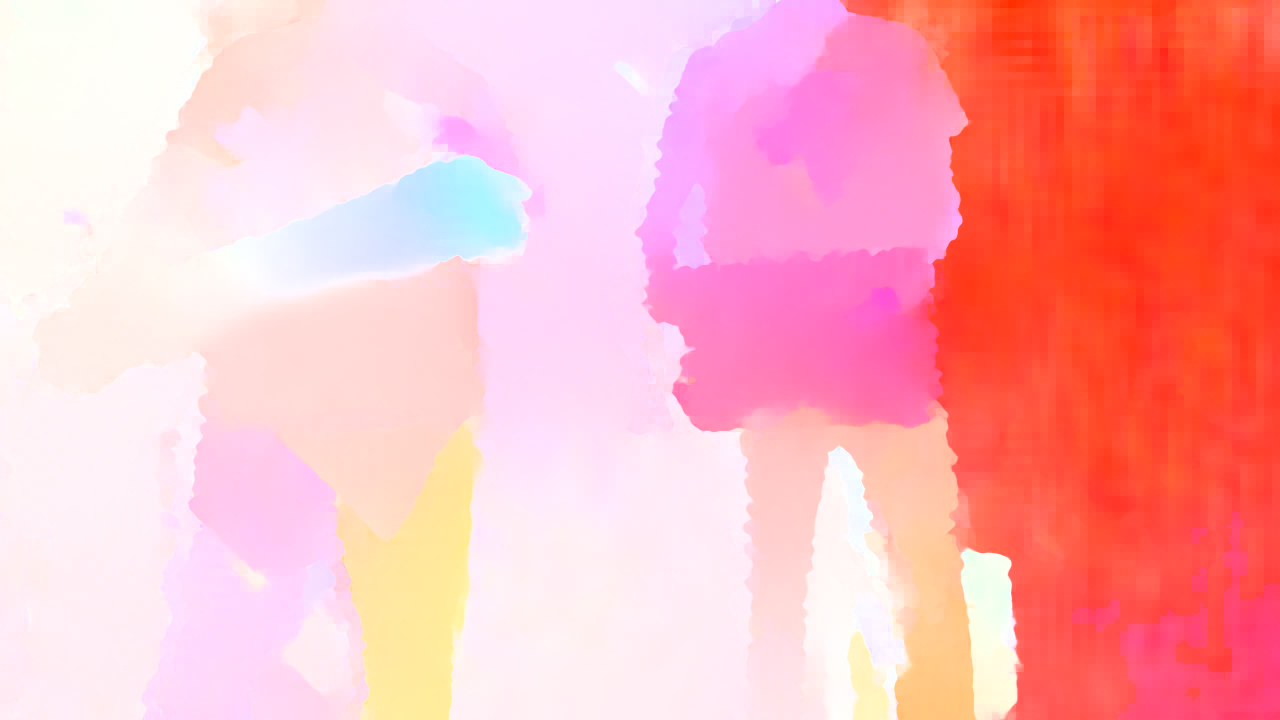}
    \end{overpic}
    \begin{overpic}[width=.12\linewidth]{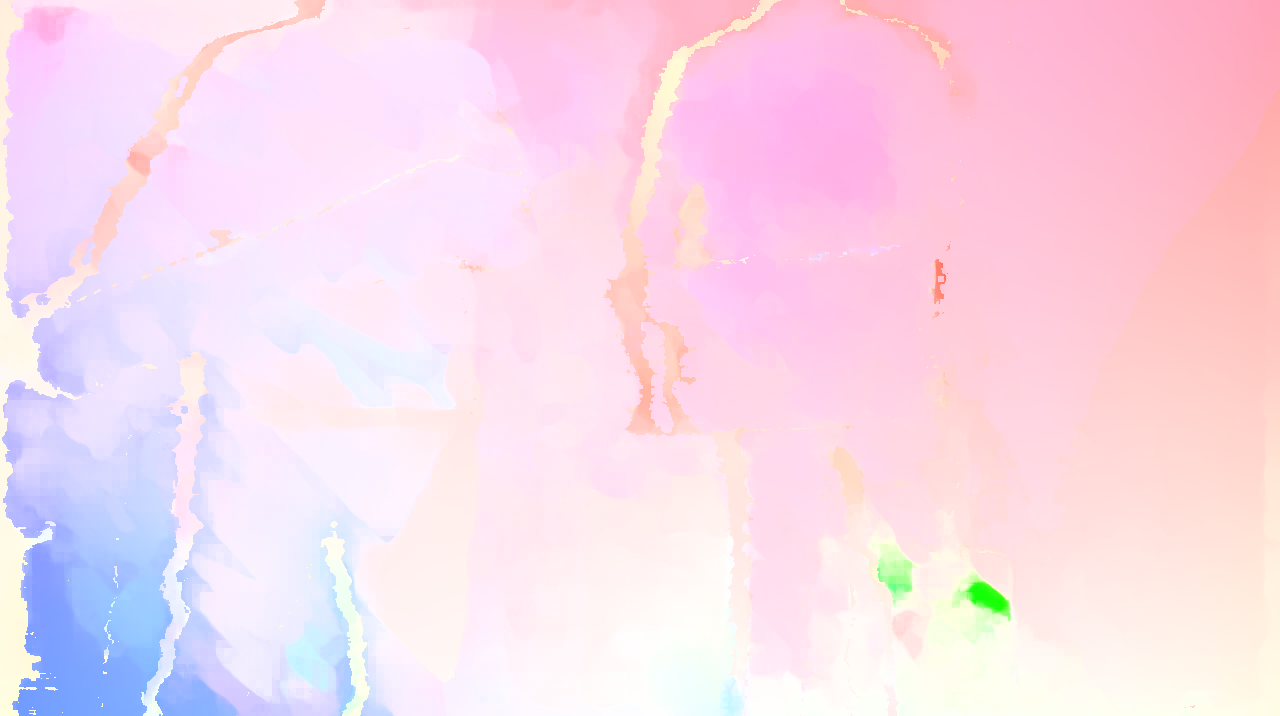}
    \end{overpic}
    \begin{overpic}[width=.12\linewidth]{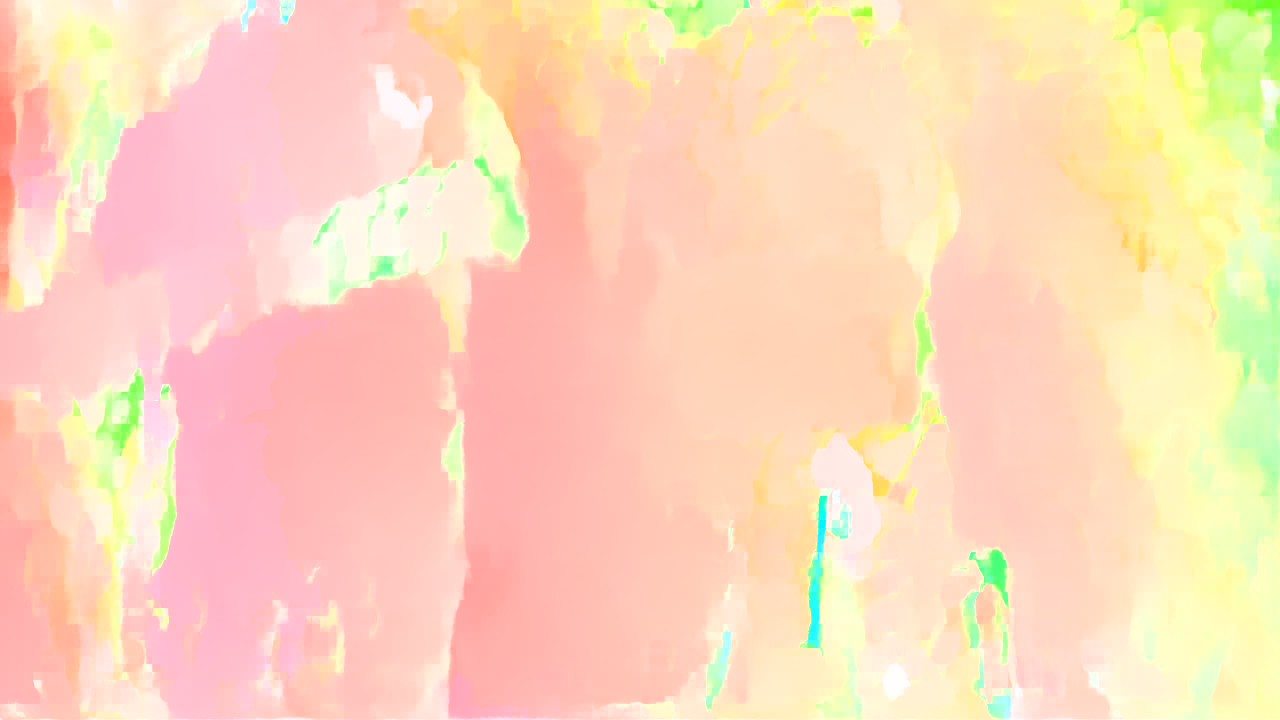}
    \end{overpic}
    \begin{overpic}[width=.12\linewidth]{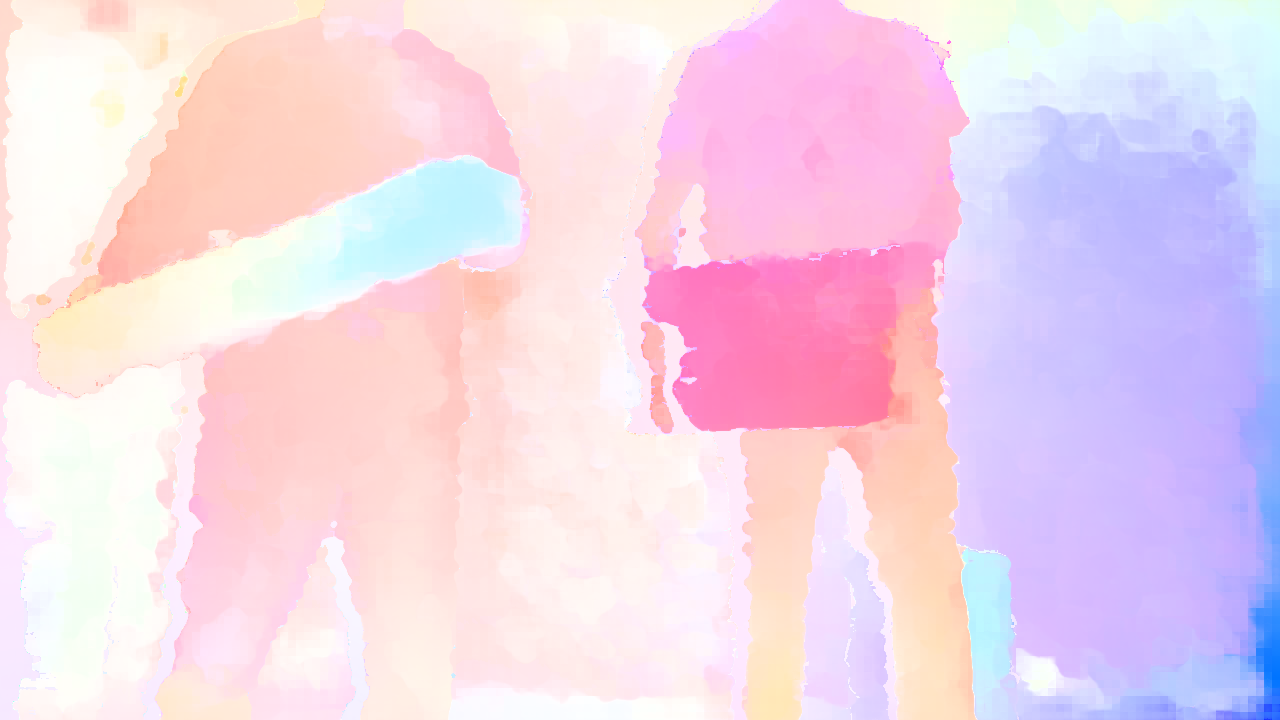}
    \end{overpic}\\
  \end{tabular}
\end{center}
\vspace{-0.2cm}
\caption{Examples of optical flow estimation error in our real-world dataset.
(top) Bright setting, (middle) Dimmed setting, (bottom) Dark setting.
{\mname} method can handle also the Dark setting (see sharper flow boundaries). Best viewed in color.}
\label{fig:realtest}
\end{figure*}

Tab.~\ref{tab:setting1_results} (middle) reports optical flow results in AGN setting.
{\mname}-2D outperforms CRAFT by $+3.94\%$ and $+3.96\%$ in terms of $\rm ACC_{1px}$, and $+1.15$ and $+1.29$ in terms of $\rm AEPE^{all}_{2D}$ in FlyingThings3D-clean and FlyingThings3D-final, respectively.
{\mname}-3D outperforms RAFT-3D by $+2.16\%$ and $+2.45\%$ in in terms of $\rm ACC_{1px}$ and $+0.51$ and $+0.61$ in terms of $\rm AEPE^{all}_{2D}$. 
$\rm AEPE^{all}_{2D}$ of RAFT-3D is lower than that of the Standard setting.
This is because $\rm AEPE^{all}_{2D}$ computes the average of all errors, and the average is known to be sensitive to outliers. 
In fact by computing the median (less sensitive to outliers), the performance of RAFT-3D in AGN setting is worse than the Standard setting: 
e.g.~RAFT-3D in FlyingThings3D-clean achieves a median $\rm EPE^{all}_{2D}$ of 0.127 and 0.130 in the Standard and AGN settings, respectively.

Tab.~\ref{tab:setting1_results} (bottom) reports optical flow results in Dark setting.
RGB methods perform worse than the previous settings, whereas our {\mname} methods outperform RGB methods, RAFT-3D, and CamLiRAFT.
Tab.~\ref{tab:setting1_results} also reports $\rm AEPE^{all}_{2D}$ and $\rm Fl^{all}_{2D}$ on KITTI without fine-tuning the models.
Although CamLiRAFT demonstrates a better generalization capability than RAFT-3D, {\mname} outperforms almost all the other methods in all three settings, demonstrating its robustness and adaptability in real-world scenarios.

Tab.~\ref{tab:4} reports scene flow results.
{\mname}-3D outperforms RAFT-3D on both FlyingThings3D-clean and FlyingThings3D-final.
CamLiRAFT scores on par with {\mname}-3D in all three settings.
{\mname}-3D performs better in terms of $\rm ACC_{0.05m}$, while CamLiRAFT performs better in terms of $\rm ACC_{0.10m}$. 
This suggests that {\mname}-3D produces more small flow errors than CamLiRAFT.
In general, {\mname}-3D performs stably across all three settings, while the performance of RAFT-3D and CamLiRAFT degrades in the AGN and Dark settings.
On KITTI, {\mname}-3D is the best-performing method on all three settings in terms of $\rm Fl^{all}_{3D}$. 
In terms of $\rm AEPE_{3D}^{all}$, CamLiRAFT performs better in the Standard and AGN settings, while {\mname}-3D scores the best in the Dark setting.

\subsubsection{Qualitative results}

We provide examples of qualitative optical flow results indicated with their corresponding $\rm AEPE^{all}_{2D}$.
We visualize the errors with respect to the ground-truth: the stronger the magenta, the higher the error.
Fig.~\ref{fig:FlyingThings3D} and Fig.~\ref{fig:KITTI} show the results of optical flow errors on FlyingThings3D and KITTI, respectively.
Both {\mname}-2D and {\mname}-3D consistently produce smaller $\rm AEPE^{all}_{2D}$ values than the other methods, which can also be visually verified with less magenta areas produced by our models.
Fig.~\ref{fig:realtest} shows the flow estimation on our acquired indoor dataset with RAFT, GMA, RAFT-3D, CamLiRAFT, and {\mname}.
In the Bright setting (top), all compared methods produce good-quality results.
In the Dimmed setting (middle), RAFT, GMA, and CamLiRAFT show low-quality results, which we can observe from the poor edges produced by the moving objects.
In the Dark setting, {\mname} is the only method that produces results where the moving objects are distinguishable.

\subsection{Ablation study}
Tab.~\ref{tab:ablation_setting1} reports the ablation study on self-attention (SA), cross-attention (CA), and Multimodal Transfer Module (MMTM) on the FlyingThings3D dataset in both Standard and Dark settings. 
Overall, we can observe that all the components we added provide an incremental contribution to improve the quality of the output optical flow compared to the RGB baseline.
SA and CA consistently improve performance (see Exp 3 vs 6 vs 8, 4 vs 7 vs 9, and similarly for the Dark setting).
The SA applied to both depth and RGB is better than applying it to the RGB branch only (see Exp 5 vs 6 for the Standard setting, and 14 vs 15 for the Dark setting).
MMTM fusion consistently outperforms the simple concatenation of RGB and depth branches in the Dark setting (see Exp 12 vs 13, 15 vs 16, 17 vs 18).
There is one case in the Standard setting where this last does not occur (see Exp 6 vs 7).
In general, SA focuses on intra-modality relationships while CA focuses on inter-modality relationships. MMTM further exchanges information across modalities at a deeper level. The best performance is achieved when all the modules are activated.

\renewcommand{\arraystretch}{.9}
\begin{table*}[!t]
    \centering
    \caption{Ablation study in Standard and Dark settings on FlyingThings3D.
    $\rm MEAN_{ACC}$ and $\rm MEAN_{AEPE}$ are the mean of $\rm ACC_{1px}$ and $\rm AEPE^{all}_{2D}$, respectively, on FlyingThings3D-clean and FlyingThings3D-final. \checkmark(RGB): self-attention is performed on the RGB.}
    \vspace{-.2cm}
    \label{tab:ablation_setting1}
    \resizebox{1\linewidth}{!}{%
    \begin{tabular}{cllcccccccccc}
        \toprule
        & \multirow{2}{*}{Exp} & \multirow{2}{*}{RGB} & \multirow{2}{*}{depth}  &  \multirow{2}{*}{SA}&  \multirow{2}{*}{CA} &  \multirow{2}{*}{Fusion} &\multicolumn{2}{c}{FlyingThings3D-clean}  &  \multicolumn{2}{c}{FlyingThings3D-final}&\multirow{2}{*}{$\rm MEAN_{ACC}$}&\multirow{2}{*}{$\rm MEAN_{AEPE}$} \\
        & & & & & & & $\rm ACC_{1px}$& $\rm AEPE^{all}_{2D}$&$\rm ACC_{1px}$&$\rm AEPE^{all}_{2D}$\\
        \midrule
        \multirow{9}{*}{\rotatebox[origin=c]{90}{Standard setting}} & 1 & \texttt \checkmark& - & - & -& - & 77.06 &4.69& 76.91&4.39&76.99 &4.54\\
        & 2 & \texttt \checkmark& - & \checkmark & - & - &77.81& 4.50 & 77.72 &4.26&77.77&4.38\\
        & 3 & \texttt \checkmark&\checkmark & - & -& concat &79.21 & 3.74&79.06 &3.71&79.14 &3.73 \\
        & 4 & \texttt \checkmark&\checkmark & - & -& MMTM &79.43& 3.79&79.27 &3.69& 79.35&3.74 \\
        & 5 & \texttt \checkmark&  \checkmark& \checkmark(RGB) & - & concat & 79.35 & 3.72 & 79.18 & 3.66& 79.27&3.69\\
        & 6 & \texttt \checkmark&  \checkmark& \checkmark & - & concat & 79.71 & 3.81 & 79.55 & 3.63& 79.63&3.72\\
        & 7 & \texttt \checkmark&  \checkmark& \checkmark & - & MMTM & 78.97 & 3.73 & 78.77 & 3.65& 78.87&3.69 \\
        & 8 & \texttt \checkmark&  \checkmark & \checkmark &  \checkmark & concat & 80.14 & 3.56 & 79.95& 3.53&80.05&3.55\\
        & 9 & \texttt \checkmark&  \checkmark & \checkmark & \checkmark & MMTM & 80.37  & 3.52 &80.21&3.42&80.29&3.47 \\
        \midrule
        \multirow{9}{*}{\rotatebox[origin=c]{90}{Dark setting}} & 10 & \texttt \checkmark& - & - & -& - & 60.26 & 8.15& 60.36& 7.85&60.31&8.00\\
        & 11 & \texttt \checkmark& - & \checkmark & - & - &67.92& 8.01 & 67.80 &7.77&67.86&7.89\\
        & 12 & \texttt \checkmark&\checkmark & - & -& concat &75.33 & 4.06&75.17 &4.05&75.25&4.06 \\
        & 13 & \texttt \checkmark&\checkmark & - & -& MMTM &75.56 & 3.97&75.40&3.99&75.48&3.98 \\
        & 14 & \texttt \checkmark&\checkmark & \checkmark(RGB) & -& concat &75.47& 3.95&75.32&3.94&75.40&3.95 \\
        & 15 & \texttt \checkmark&  \checkmark& \checkmark & - & concat & 75.69 & 3.96 & 75.57 & 3.90&75.63&3.93\\
        & 16 & \texttt \checkmark&  \checkmark & \checkmark & - & MMTM & 75.75 & 3.81 & 75.60& 3.76&75.68&3.79\\
        & 17 & \texttt \checkmark&  \checkmark & \checkmark & \checkmark & concat &76.43  & 3.79 &76.26 &3.72&76.35&3.76 \\
        & 18 & \texttt \checkmark&  \checkmark & \checkmark & \checkmark & MMTM & 76.70  & 3.65 &76.57 &3.66&76.64&3.66 \\
        \bottomrule 
    \end{tabular}
    }
\end{table*}
\renewcommand{\arraystretch}{1}

\subsection{Computation analysis}
We measure the number of parameters, Floating-Point Operations (FLOPs), and inference time of all compared methods using FlyingThings3D.
We conducted the experiments with a Nvidia 3090 GPU (24G) and I9-10900 CPUs, and reported the results in Tab.~\ref{tab:cost}.
Despite {\mname}-2D has the second-largest number of parameters, its number of FLOPs and inference time are in-between the other methods for optical flow estimation. The inference time of {\mname}-3D is slightly higher than that of CamLiRAFT, although our number of parameters is one order of magnitude larger than CamLiRAFT.
From the per-component analysis of {\mname}-2D in Tab.~\ref{tab:costabla}, we can observe that Self-attention and Cross-attention have a higher computational cost than MMTM and the two-branch encoder. The most time-consuming component is \textit{Others} which includes all the other modules to compute the optical flow.

\renewcommand{\arraystretch}{.9}
\begin{table}[t]
    \centering
    \caption{Comparative computational analysis by using 960$\times$540-sized images on a Nvidia 3090.}
    \vspace{-.3cm}
    \label{tab:cost}  
    \resizebox{1\linewidth}{!}{%
    \begin{tabular}{lccc}
        \toprule
        Models & Params [M] & FLOPs [T] & Inference time [ms] \\
        \midrule
        RAFT \cite{Raft} & 5.31 & 0.78 & 99\\
        GMA \cite{GMA} & 5.88 & 0.59 & 87 \\
        Separable flow \cite{Separable_flow} & 8.35 & 0.50 & 639 \\
        CRAFT \cite{CRaft} & 6.31 & 0.99 & 302 \\
        {\mname}-2D& 8.13 & 0.68 & 219\\
        \midrule
        RAFT-3D \cite{Raft-3D} & 44.50 & 0.51 & 277 \\
        CamLiRAFT \cite{camliraft2023} & 8.41 & 0.67 & 312 \\
        {\mname}-3D & 86.32 & 0.76 & 398 \\  
        \bottomrule 
    \end{tabular}
    }
\end{table}

\begin{table}[t]
    \centering
    \caption{Ablations of computational performance on {\mname}-2D by using 960$\times$540-sized images on a Nvidia 3090.}
    \vspace{-.3cm}
    \label{tab:costabla}  
    \resizebox{1\linewidth}{!}{%
    \begin{tabular}{lccc}
        \toprule
        Models & Params [M] & FLOPs [T] & Inference time [ms] \\
        \midrule
        Image encoder & 2.33 & 0.11 & 13\\
        Depth encoder & 2.33 & 0.11 & 13\\
        Self-attention & 0.16 & 0.06 & 63 \\
        Cross-attention & 0.09 & 0.04 & 43 \\
        MMTM & 0.24 & 2M &1\\
        Others & 2.98 & 0.36 & 86\\        
        \midrule
        Total & 8.13 & 0.68 & 219\\
        \bottomrule 
    \end{tabular}
    }
\end{table}
\renewcommand{\arraystretch}{1}

\section{Conclusions}

We presented \mname, a novel approach for optical and scene flow estimation.
{\mname} improves feature extraction with an early-fusion Multimodal Feature Fusion (MFF) Encoder. 
MFF attends to informative features and enables information exchange within and across modalities by using self-attention, cross-attention, and the Multimodal Transfer Module.
Through experimental validation, we showed that {\mname} generates more stable and informative feature descriptions by exploiting the different modalities.
{\mname} scores state-of-the-art results in Standard setting, but also in our newly introduced AGN and Dark settings where RGB information is corrupted.
Future research directions may include the integration of {\mname} in robotic systems for autonomous navigation.


\IEEEpubidadjcol

\bibliographystyle{IEEEtran}
\bibliography{refe}

\end{document}